\documentclass{article}
\usepackage{iclr2027_conference,times}

\usepackage[utf8]{inputenc}
\usepackage[T1]{fontenc}
\usepackage{hyperref}
\usepackage{url}
\usepackage{booktabs}
\usepackage{amsfonts}
\usepackage{amsmath}
\usepackage{amssymb}
\usepackage{nicefrac}
\usepackage{microtype}
\usepackage{xcolor}
\usepackage{colortbl}
\usepackage{graphicx}
\usepackage{placeins}
\graphicspath{{figures/}}

\title{Flattening the Connectome Spectrum:\\ A Spectral Filter for FC Induces a Pretraining Target for fMRI Encoders}
\author{Giovanni Marraffini$^{1,2}$, Victoria Shevchenko$^{2}$,\\
\textbf{Carlo Alberto Barbano$^{1}$, Demian Wassermann$^{1}$}\\
$^{1}$Inria Saclay \^Ile-de-France, CEA, Universit\'e Paris-Saclay, Palaiseau, France\\
$^{2}$Sigma Nova\\
\texttt{giovanni.marraffini@gmail.com}}

\iclrfinalcopy

\begin{document}

\maketitle
\lhead{Preprint}

\begin{abstract}
Self-supervised pretraining reshaped prediction in language and vision, and brain foundation models (BFMs) inherited its promise. Representations learned from large unlabelled corpora should capture individual functional dynamics and generalise across cohorts. However, kernel ridge regression (KRR) fitted on functional connectivity (FC) matrices still predicts individual phenotypes more accurately than any BFM we tested. In this paper, we show that KRR is weighted by the eigenvalues of the FC which are miscalibrated for phenotype prediction. We apply an efficient spectral filter to recalibrate the eigenvalues of each subject's FC matrix, enabling the model to exploit more inter-individual variance. Across the $5$ datasets, $11$ parcellations and $6$ prediction targets we tested, we match or exceed the KRR baseline. Based on this finding, we then pretrain a small encoder model on about $4{,}000$ hours of fMRI from $162$ open datasets, whereby we align the pairwise similarities between the embeddings of recording snippets with those between the recalibrated connectomes. Our model performs on par with the best of the $6$ published BFMs we tested while having an order of magnitude fewer parameters. Our encoder performs better than FC on short scans and in smaller cohorts, especially in fingerprinting. We release the pretrained model weights, the code and the pretraining data, preprocessed and parcellated.
\end{abstract}

\section{Introduction}

Functional connectivity (FC), the matrix of pairwise correlations between regional fMRI timeseries, reliably captures subject-specific variance, enabling the prediction of subject identity from distinct scanning sessions \citep{finn2015functional}. FC also allows us to predict cognitive performance measures such as fluid and crystallized intelligence \citep{kong2023comparison, ooi2022comparison}. This capability is likely to stem from FC variance being dominated by stable individual and group factors rather than by the mental or physical state during the scanning session \citep{gratton2018functional}. \citet{ooi2022comparison} built a prediction pipeline on these matrices: kernel ridge regression on the off-diagonal FC entries predicts cognitive ability and a wide range of behavioural measures.

In recent years, Brain Foundation Models (BFMs) pretrained on fMRI time-series brought the promise of surpassing such linear baselines in prediction from fMRI, and of representations that transfer across cohorts and tasks. Nevertheless, BFMs are currently underperforming simple linear estimators fitted on FC \citep{he2020deep, marraffini2026variance}. Most common explanations for this disparity are centered on data scarcity \citep{Marek2022-zh} and low signal-to-noise ratio (SNR) \citep{Schulz2024-la}. In particular, for low SNR data such as fMRI, the adequacy of the default masked-reconstruction pretraining objective has been questioned \citep{Van-Assel2025-nl}. All in all, they yield representations that do not capture as much behaviorally relevant information as FC \citep{zhou2025brain}. Therefore, instead of simply assuming these models should scale to beat these baselines, we first study where the linear estimator on FC itself fails and we improve it. Then, leveraging this transformation, we craft a pretraining loss that makes a small model trained on open data match the best published BFMs we tested and beat KRR on short scans and fewer subjects.

\begin{figure}[!t]
\centering
\includegraphics[width=\linewidth]{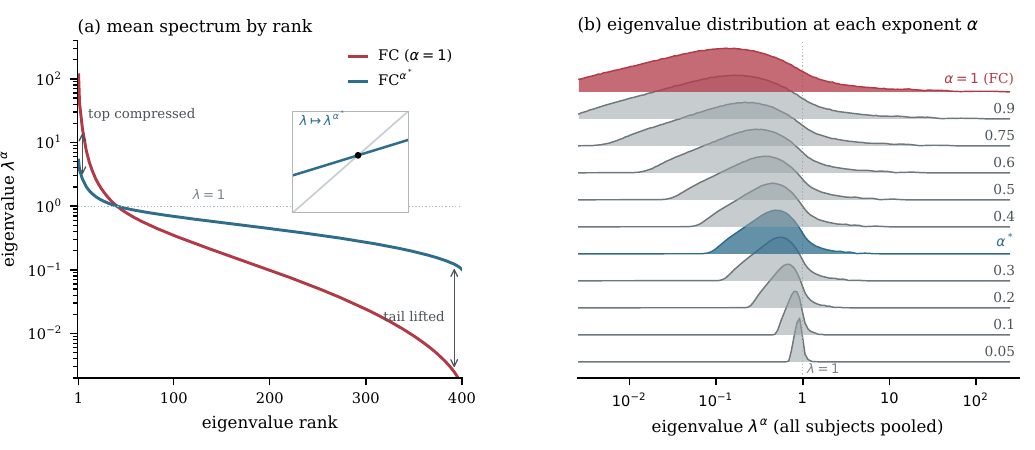}
\caption{\textbf{(a)} The eigenvalue spectrum of the FCs from HCP-YA \citep{VANESSEN201362}  by rank, averaged over the $955$ subjects, before ($\alpha=1$, red) and after raising the eigenvalues to $\alpha^{*}$ (blue). The transform is concave with a fixed point at $\lambda=1$, so eigenvalues are pushed onto an increasingly compact support around 1, with their order preserved. \textbf{(b)} The distribution of eigenvalues pooled over the same $955$ subjects, from $\alpha=1$ (red, raw FC) down to $\alpha=0.05$; the one at $\alpha^{*}$ is blue. Each is the histogram of $\log_{10}\lambda^{\alpha}$; lowering $\alpha$ pulls the distribution toward $\lambda=1$ from both sides (Appendix~\ref{app:sphere}).}
\label{fig:powermap}
\end{figure}

We find that an answer lies in the spectrum of the FC matrix itself. We observe that the space spanned by a small percentage of the eigenvectors of a subject's FC matrix carries almost all of the weight the KR regression uses, while the variance that distinguishes individuals is spread over a larger space (Figure~\ref{fig:powermap}, Table~\ref{tab:matched}). We also observe that flattening the FC's spectrum by applying an exponentiation between 0 and 1 improves the prediction on most phenotypes on every dataset and parcellation we test. In particular, it improves fingerprinting far more than sex or age prediction (Table~\ref{tab:bench}). The same spectrum-flattened connectome serves as a pretraining target. We train a small encoder to preserve the pairwise similarities between the spectrum-flattened connectomes which then reaches comparable performance to the best published BFMs we tested on most datasets and tasks, with an order of magnitude fewer parameters. Furthermore, in some tasks and datasets it gets significantly better than every other tested BFM. In summary, our contributions are:

\begin{itemize}\itemsep1pt
\item \textbf{Recovering individual variance from FC with a spectral transformation.} We find that the eigenvalue distribution of each subject FC is extremely uneven. We derive how this concentrates the regression on a few percent of the eigenvectors (Section~\ref{sec:transform}, Table~\ref{tab:matched}, Appendix~\ref{app:mechanism}). We recalibrate the matrix at the selected exponent $\alpha^{*}$ ($\alpha^{*} = 0.35$ throughout this paper, chosen in Section~\ref{sec:alpha}), which significantly improves prediction (Tables~\ref{tab:main} and \ref{tab:bench}).

\item \textbf{A pretraining objective derived from the transformation.} Using pairwise similarities between $\mathrm{FC}^{\alpha^{*}}$ matrices as a pretraining target, an encoder learns to match these similarities in the encoded embeddings from only short windows of the full scan. \textbf{This encoder matches or exceeds the best foundation models we tested on almost every dataset and task, with roughly an order of magnitude fewer parameters and trained fully on open and free data}. In particular, given short scans and fewer subjects, the encoder outperforms FC on the cognitive composite and in fingerprinting (Tables~\ref{tab:bench} and \ref{tab:grid}). The objective applies to any multivariate timeseries for which a pairwise similarity between recordings can be computed.

\end{itemize}

We release the code, the encoders with $1$M and $8$M parameters and the pretraining data as used: $162$ open datasets, $9{,}578$ subjects, $27{,}835$ recordings, preprocessed and parcellated.\footnote{\raggedright Code: \url{https://github.com/GioMarraffini/fc-power}\newline Encoders: \url{https://huggingface.co/Marraffini-Giovanni/fbert-1m}\newline \phantom{Encoders: }\url{https://huggingface.co/Marraffini-Giovanni/fbert-8m}\newline Data: \url{https://huggingface.co/datasets/Marraffini-Giovanni/fbert-corpus}}

\section{Related work}
\label{sec:related}

\paragraph{Other transforms of the connectome.} Predicting from raw correlation matrices has been shown to be suboptimal: the tangent-space parameterisation \citep{varoquaux2010detection}, a matrix logarithm after whitening by a group reference, outperforms raw correlations in connectome-prediction benchmarks \citep{dadi2019benchmarking, pervaiz2020optimising}. The power-Euclidean family $\mathbf{\Sigma}\mapsto\mathbf{\Sigma}^{\alpha}$ \citep{dryden2010power} contains the raw matrix at $\alpha=1$ and the log map as $\alpha\to0$, since $(\mathbf{\Sigma}^{\alpha}-\mathbf{I})/\alpha\to\log\mathbf{\Sigma}$ (Appendix~\ref{app:geodesic}). Where in this family the optimum for behavioural prediction lies has not been reported; Appendix~\ref{app:screen} shows that the extremes are worse than mid values.

\paragraph{Predicting phenotypes from FC.} \citet{ooi2022comparison} compare regression methods and feature sets for behavioural prediction on HCP-YA and define the protocol we adopt unchanged: kernel ridge regression with a correlation kernel on the off-diagonal entries, replicated nested cross-validation with family-aware splits, and a broad battery of behavioural targets. \citet{kong2023comparison} and \citet{dubois2018distributed} adopt comparable pipelines, and deep networks do not beat this regression \citep{he2020deep}. We leave the regression, the kernel and the cross-validation as they are and change only the matrix that enters them.

\paragraph{Brain foundation models.} Recent rs-fMRI BFMs share a Transformer or state-space backbone and differ in the pretraining objective. BrainLM \citep{caro2023brainlm} is a ViT-MAE trained by masked reconstruction; Brain-JEPA \citep{dong2024brain} predicts masked-patch representations in embedding space; BrainMass \citep{yang2024brainmass} takes FC matrices rather than timeseries and combines masked ROI modelling with latent representation alignment; BrainHarmonix \citep{dong2025brainharmonix} adds structural morphology to a functional encoder. Brain-Semantoks \citep{gijsen2025brainsemantoks} learns semantic tokens of the timeseries with a self-distilled objective. Standardised comparison against FC baselines remains rare \citep{zhou2025brain}; Section~\ref{sec:bench} provides one.

\paragraph{Distillation with a similarity target.} Matching a student's pairwise similarities to a teacher's is an established form of distillation \citep{fang2021seed, park2019relational, tung2019similarity, tian2020contrastive}. Grading pairs by a kernel instead of a hard positive/negative split is the mechanism of the kernel contrastive loss of \citet{barbano2023contrastive}, which sets a degree of positiveness from a continuous label. Centred kernel alignment \citep{cortes2012algorithms}, also known as CKA \citep{kornblith2019similarity}, is the normalised similarity of two Gram matrices. In all of these the teacher is a trained network or a label; Section~\ref{sec:distill} uses a closed-form statistic of the full recording instead.

\section{Methods}
\label{sec:methods}

\subsection{Predicting from the connectome}
\label{sec:regression}

\textbf{Correlation-kernel KRR.} We follow the protocol of \citet{ooi2022comparison}: kernel ridge regression with a Pearson correlation kernel on the off-diagonal entries of the matrix, the ridge penalty selected by inner cross-validation (Appendix~\ref{app:protocol}).

\textbf{Linear-kernel ridge probe (Table~\ref{tab:bench}).} For the comparison with BFMs (Table~\ref{tab:bench}) we use the same regression with a linear kernel on z-scored features (Appendix~\ref{app:protocol}). Both are kernel ridge regression and differ only in the kernel: the correlation kernel normalises each subject's feature vector, but a BFM may carry information in that norm, which the correlation kernel discards. The two agree to within $0.02$ (Appendix~\ref{app:protocol}).

\subsection{The transform}
\label{sec:transform}

For each subject, let $\mathbf{Z}\in\mathbb{R}^{P\times T}$ be the z-scored parcel timeseries and $\mathbf{\Sigma}=\frac{1}{T-1}\mathbf{Z}\mathbf{Z}^{\top}$ its Pearson FC matrix. $\mathbf{\Sigma}$ is symmetric positive definite, so $\mathbf{\Sigma}=\mathbf{V}\mathbf{D}\mathbf{V}^{\top}$ with orthonormal eigenvectors $\mathbf{v}_1,\dots,\mathbf{v}_P$ and $\mathbf{D}=\mathrm{diag}(\lambda_1\ge\dots\ge\lambda_P>0)$. The transform is
\begin{equation}
\mathbf{\Sigma}^{\alpha} \;=\; \mathbf{V}\,\mathbf{D}^{\alpha}\,\mathbf{V}^{\top},\qquad \mathbf{D}^{\alpha}=\mathrm{diag}(\lambda_1^{\alpha},\dots,\lambda_P^{\alpha}).
\end{equation}
It is computed per subject, with no group mean, and no parameter estimated from other subjects.

\textbf{Why the eigenvalues matter to the regression.} The regression of Section~\ref{sec:regression} vectorises the matrix and applies a correlation kernel, so the similarity between subjects $a$ and $b$ is the Frobenius inner product of their matrices up to centring and scaling. For eigendecompositions $\mathbf{\Sigma}_a^{\alpha}=\sum_i\lambda_i^{\alpha}\mathbf{v}_i\mathbf{v}_i^{\top}$ and $\mathbf{\Sigma}_b^{\alpha}=\sum_j\mu_j^{\alpha}\mathbf{u}_j\mathbf{u}_j^{\top}$,
\begin{equation}
\big\langle \mathbf{\Sigma}_a^{\alpha},\mathbf{\Sigma}_b^{\alpha}\big\rangle_F
\;=\;\sum_{i,j}\lambda_i^{\alpha}\,\mu_j^{\alpha}\,(\mathbf{v}_i^{\top}\mathbf{u}_j)^2 .
\label{eq:kernel}
\end{equation}
The kernel is a weighted sum of squared overlaps between the two subjects' eigenvectors, and the weight on the overlap of mode $i$ with mode $j$ is $\lambda_i^{\alpha}\mu_j^{\alpha}$. Over $n$ subjects, with $\tilde{\mathbf{\Sigma}}_a^{\alpha}$ the matrix $\mathbf{\Sigma}_a^{\alpha}$ centered, the method consumes the $n\times n$ correlation kernel $\mathbf{K}_{\alpha}$ 
\begin{equation}
[\mathbf{K}_{\alpha}]_{ab} \;=\; \frac{\big\langle \tilde{\mathbf{\Sigma}}_a^{\alpha},\tilde{\mathbf{\Sigma}}_b^{\alpha}\big\rangle_F}{\|\tilde{\mathbf{\Sigma}}_a^{\alpha}\|_F\,\|\tilde{\mathbf{\Sigma}}_b^{\alpha}\|_F}
\;=\;[\mathbf{K}_{\alpha}]_{ba},
\label{eq:Kalpha}
\end{equation}
which is symmetric and positive semidefinite with unit diagonal, and we write $\bar{\mathbf{K}}_{\alpha}=\mathbf{H}\mathbf{K}_{\alpha}\mathbf{H}^{\top}$ for its doubly centred form, with $\mathbf{H}=\mathbf{I}-\tfrac{1}{n}\mathbf{1}\mathbf{1}^{\top}$. Kernel regression fits the components of the target that lie along the leading eigendirections of its kernel first and most strongly \citep{cao2021spectral, tancik2020fourier}, so a kernel this concentrated can use only a few directions of variation between subjects, and reshaping the kernel spectrum is a direct way to change this inductive bias \citep{geifman2023controlling}. The exponent does it at the level of each subject's matrix, before the kernel is formed, and changes only these weights. Across the HCP-YA subjects, $\alpha^{*}$ moves the participation ratio $(\sum_i\lambda_i)^2/\sum_i\lambda_i^2$ from $10.1\pm4.1$ to $218.9\pm18.2$.

\begin{figure}[!t]
\centering
\includegraphics[width=\linewidth]{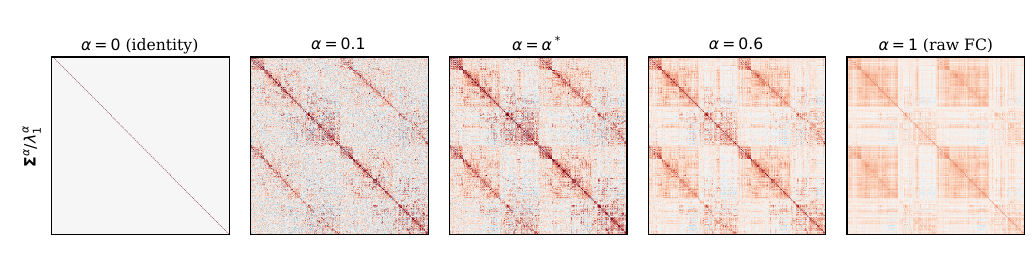}
\caption{The recalibrated connectome $\mathbf{\Sigma}^{\alpha}$ of one HCP-YA subject at Schaefer-400 for $\alpha=0$, $0.1$, $\alpha^{*}$, $0.6$ and $1$, each scaled by its largest eigenvalue so that one colour scale serves the row; $\alpha=1$ is raw FC and $\alpha=0$ the identity.}
\label{fig:heat}
\end{figure}

\textbf{Interpretation.} With the point of view of signal processing on graphs \citep{shuman2013emerging}, $\mathbf{\Sigma}^{\alpha}$ applies the spectral filter $h(\lambda)=\lambda^{\alpha}$ to the eigenmodes of the connectome: for $\alpha<1$ it is a monotone equaliser that compresses the dynamic range of the modes without reordering them, from a ratio $\lambda_1/\lambda_P$ (between the largest and the smallest eigenvalues) of about $10^{5}$ at $\alpha=1$ to about $50$ at $\alpha^{*}$, and at $\alpha=0$ it is a whitening filter that maps every subject to the identity. Figure~\ref{fig:heat} shows the filtered connectome of one subject along this path. Appendix~\ref{app:alpha} carries this interpretation in full and explores two others.

\subsection{Choosing the exponent}
\label{sec:alpha}

We use two different methods to find $\alpha^{*}$. First, we find it using the nested cross-validated prediction of the cognitive composite under the correlation-kernel KRR \footnote{For this grid search, we used HCP-YA at Schaefer-400}. We do it over a grid of possible alphas (Appendix \ref{app:alpha-choice}), with each subject's eigendecomposition computed once and reused so that every exponent is scored on identical folds (Figure~\ref{fig:alpha}a). We find $\alpha^{*}=0.35$ to be optimal, and every result in the paper uses this value. 

Second, we search for the optimal exponent $\alpha$ so that the pairwise differences between the $\mathrm{FC}^{\alpha^{*}}$ matrices are as similar as possible to the pairwise differences between the individual traits. This is equivalent to searching for the optimal $\alpha$ to maximize the centered kernel-target alignment \citep{cortes2012algorithms} between the kernel $\bar{\mathbf{K}}_{\alpha}$ of Section~\ref{sec:transform} over the training subjects and the target vector (individual features) $\mathbf{y}$. This metric,
\begin{equation}
\mathrm{A}(\alpha)=\frac{\langle\bar{\mathbf{K}}_{\alpha},\bar{\mathbf{Y}}\rangle_F}{\|\bar{\mathbf{K}}_{\alpha}\|_F\,\|\bar{\mathbf{Y}}\|_F},\qquad
\bar{\mathbf{Y}}=\mathbf{H}\mathbf{y}\mathbf{y}^{\top}\mathbf{H},\qquad
\alpha_{\mathrm{A}}=\arg\max_{\alpha\in(0,1]}\mathrm{A}(\alpha),
\label{eq:kta}
\end{equation}
measures how well the kernel's geometry matches the target. It is smooth and its derivative is available in closed form. We maximize $\mathrm{A}$ with the method by \citet[Appendix~\ref{app:alpha-choice}]{brent1973algorithms}. It gives $\alpha_{\mathrm{A}}=0.387$, and using $\alpha^{*}$, we stay at $99.55\%$ of its maximum (Figure~\ref{fig:alpha}b). The same alignment between two kernels, with a teacher kernel in place of $\bar{\mathbf{Y}}$, is the pretraining objective of Section~\ref{sec:distill}.

\subsection{Pretraining by distillation}
\label{sec:distill}

We pretrain the encoder in a student-teacher distillation manner by maximizing the alignment between the student (encoder) and the teacher kernels.

\begin{figure}[!t]
\centering
\includegraphics[width=0.96\linewidth]{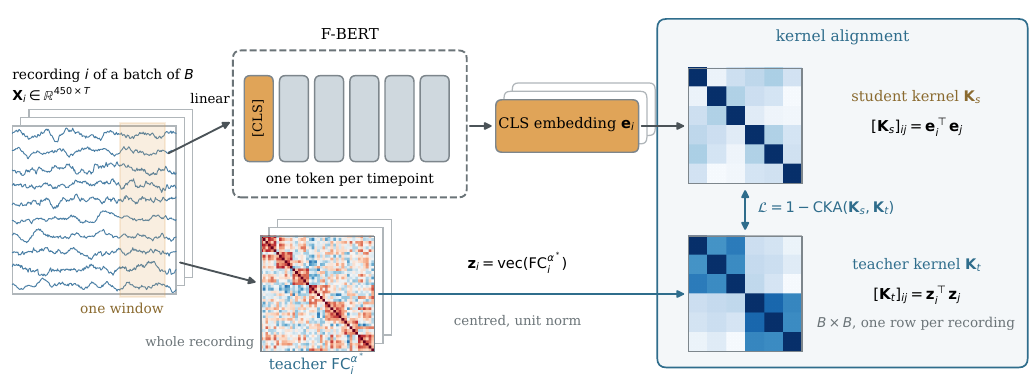}
\caption{The kernel-alignment objective. Each timepoint of the window becomes one token through a linear projection, and a prepended \texttt{[CLS]} token carries the loss. The teacher is the $\mathrm{FC}^{\alpha^{*}}$ of the entire recording, vectorised, centred and scaled to unit norm. Every recording of a batch of $B$ goes through both lanes, and the $B$ embeddings and the $B$ teacher vectors each define a $B\times B$ kernel, $[\mathbf{K}_s]_{ij}=\mathbf{e}_i^{\top}\mathbf{e}_j$ and $[\mathbf{K}_t]_{ij}=\mathbf{z}_i^{\top}\mathbf{z}_j$, and the loss is one minus their centred alignment (Equation~\ref{eq:cka}).}
\label{fig:distill}
\end{figure}

\textbf{Student.} The encoder is fMRI-BERT (Figure~\ref{fig:distill}), a bidirectional Transformer encoder with the BERT architecture \citep{devlin2019bert} over the parcellated timeseries: each timepoint of a $P$-region sequence is projected linearly to the hidden width to form one token, a learned \texttt{[CLS]} token is prepended, and the \texttt{[CLS]} output is the embedding. Positional encoding is sinusoidal, so an encoder pretrained on short windows reads a recording of any length; at inference each recording is encoded as one continuous sequence.

\textbf{Teacher.} For each recording $i$ in a batch, $\mathbf{z}_i=\mathrm{vec}(\mathrm{FC}^{\alpha^{*}}_i)$ is the vector of off-diagonal edges of $\mathrm{FC}^{\alpha^{*}}$ computed from the whole recording, centered on its own mean and scaled to unit norm. The teacher is always computed from the full recording while the student sees only a short window of it during pretraining, so the student is trained to infer whole-recording connectivity from a partial view. This is key when trying to beat KRR on FC in short scans.

\textbf{Objective.} Let $\mathbf{E}\in\mathbb{R}^{B\times d}$ hold the \texttt{[CLS]} embeddings of a batch of $B$ windows from $B$ recordings and $\mathbf{Z}\in\mathbb{R}^{B\times P(P-1)/2}$ the corresponding teacher vectors. Their Gram matrices $\mathbf{K}_s=\mathbf{E}\mathbf{E}^{\top}$ and $\mathbf{K}_t=\mathbf{Z}\mathbf{Z}^{\top}$ are the student's and the teacher's similarity between every pair of recordings in the batch; because the teacher vectors are centered and unit-norm, $\mathbf{K}_t$ is the correlation kernel of Section~\ref{sec:regression} restricted to the batch, the similarity the regression consumes. With $\mathbf{H}=\mathbf{I}-\tfrac{1}{B}\mathbf{1}\mathbf{1}^{\top}$ the centering matrix and $\bar{\mathbf{K}}=\mathbf{H}\mathbf{K}\mathbf{H}$ a doubly centered kernel, the loss is one minus their centred alignment,
\begin{equation}
\mathcal{L}=1-\mathrm{CKA}(\mathbf{K}_s,\mathbf{K}_t),\qquad
\mathrm{CKA}(\mathbf{K}_s,\mathbf{K}_t)=\frac{\langle\bar{\mathbf{K}}_s,\bar{\mathbf{K}}_t\rangle_F}{\|\bar{\mathbf{K}}_s\|_F\,\|\bar{\mathbf{K}}_t\|_F},
\label{eq:cka}
\end{equation}
which is Equation~\ref{eq:kta} with the teacher kernel in place of the target outer product: the exponent is chosen by aligning the kernel to the phenotype, and the student is trained by aligning its kernel to the teacher's. The loss is invariant to isotropic scaling and to rotations of either representation and, like any CKA, is dominated by the directions of largest variance in each representation \citep{davari2023reliability, ding2021grounding}, which on the teacher side is exactly what the exponent flattened, and is estimated better from a larger batch. No second view of the same recording is used in the same batch, and unlike the distillation objectives of Section~\ref{sec:related} the teacher is not a network but a closed-form statistic the student cannot compute from its window.

\section{Experimental Setup} 

\subsection{Datasets}
\label{sec:data}

\textbf{HCP-YA} \citep{VANESSEN201362}. 
\textbf{AOMIC-ID1000} \citep{snoek2021amsterdam}.
\textbf{ABIDE-I} \citep{dimartino2014abide}.
\textbf{ADHD-200} \citep{adhd2012consortium}. 
\textbf{CoRR} \citep{zuo2014corr}. 
\textbf{Pretraining corpus}: $162$ openly available OpenNeuro datasets. Parcellations, preprocessing and per-dataset details are in Appendix~\ref{app:data}.

\textbf{Targets.} The primary target is a cognitive composite: on HCP-YA the first principal component of the $58$ behavioural measures of \citet{ooi2022comparison}; on AOMIC the first principal component of the four Intelligence Structure Test subscales. We also report sex and age on HCP-YA and AOMIC, autism diagnosis on ABIDE-I, ADHD diagnosis on ADHD-200 and identification on CoRR.

\subsection{Evaluation}
\label{sec:eval}

\textbf{Cross-validation.} Every score comes from a nested cross-validation with $10$ outer folds, repeated with different fold assignments ($20$ repetitions unless stated otherwise), family-aware on HCP-YA (twins and siblings never span a split), subject-level on AOMIC and site-grouped on ABIDE-I and ADHD-200 (each site held out in turn). The ridge penalty and the PCA that defines each cognitive composite are fit on training folds only (Appendix~\ref{app:protocol}).

\textbf{Statistics.} We report the mean Pearson $r$ (or AUC) and the standard deviation across folds; in Table~\ref{tab:bench} the point value is the $r$ of the pooled out-of-fold predictions. Paired comparisons between two representations use the same folds and the corrected resampled $t$-test of \citet{nadeau2003inference}, written $p_{\mathrm{NB}}$, which accounts for the dependence between folds that share training subjects.

\subsection{Pretraining}
\label{sec:pretrain-setup}

Two sizes of fMRI-BERT are trained, $128$-dimensional with $4$ layers ($1$M parameters) and $384$-dimensional with $8$ layers ($8$M), at Sch-450 (Schaefer-400 plus the Tian-S3 subcortex, $P=450$, $101{,}025$ edges). Pretraining windows are $80$ timepoints, the batch is $B=1024$ recordings, and each model is trained with AdamW under a warm-up and cosine schedule for $300{,}000$ steps; the learning rates, checkpoint frequency and the checkpoint-selection rule with its measured optimism are in Appendix~\ref{app:pretrain}. All reported scores are frozen linear probes. A KL form of the same target (Appendix~\ref{app:pretrain}) scores lower at every model size and batch size, whereas the CKA loss improves with batch size.

\section{Results}

\subsection{The transform improves prediction on every cohort tested}
\label{sec:main}

Table~\ref{tab:main} compares $\mathrm{FC}^{\alpha^{*}}$ with raw FC on the same subjects and folds under the \citet{ooi2022comparison} method: the improvement is significant on both cohorts at every parcellation (Appendix~\ref{app:table1full}). $\mathrm{FC}^{\alpha^{*}}$ also predicts better than other standard transformations of FC such as the tangent-space parameterisation, the log-Euclidean map and partial correlation (Appendix~\ref{app:screen}). The $\alpha\to0$ limit of the family scores below the peak (Figure~\ref{fig:alpha}). Appendix~\ref{app:mechanism} shows that none of the other spectral filters we tried beats $\lambda^{\alpha^{*}}$.

\begin{table}[!t]
\centering
\small
\caption{$\mathrm{FC}^{\alpha^{*}}$ against raw FC on the same folds, correlation-kernel KRR. Pearson $r$, mean $\pm$ std over folds; $\Delta$ is the paired difference and $^{*}$ marks $p_{\mathrm{NB}}<0.05$ (Appendix~\ref{app:table1full}).}
\label{tab:main}
\begin{tabular}{lccc}
\toprule
parcellation & FC ($\alpha{=}1$) & $\mathrm{FC}^{\alpha^{*}}$ & $\Delta$ \\
\midrule
\multicolumn{4}{l}{\emph{HCP-YA, resting state}} \\
Schaefer-100 & $0.505 \pm 0.088$ & $0.575 \pm 0.081^{*}$ & $+0.070$ \\
Schaefer-400 & $0.543 \pm 0.083$ & $\mathbf{0.627 \pm 0.080}^{*}$ & $+0.084$ \\
\midrule
\multicolumn{4}{l}{\emph{AOMIC-ID1000, movie watching}} \\
Schaefer-100 & $0.310 \pm 0.091$ & $0.409 \pm 0.083^{*}$ & $+0.099$ \\
Schaefer-400 & $0.348 \pm 0.084$ & $\mathbf{0.432 \pm 0.081}^{*}$ & $+0.084$ \\
\bottomrule
\end{tabular}
\end{table}

Table~\ref{tab:matched} scores the same top-$k$ eigenvectors with raw and with flattened eigenvalues. The participation ratio of raw FC is $10.1\pm4.1$ (Section~\ref{sec:transform}), and raw FC gains nothing beyond $20$ modes: $0.544$ at $20$, $0.542$ at $400$. With $\lambda^{\alpha^{*}}$ the same $20$ eigenvectors give $0.610$ and all $400$ give $0.624$. The eigenvectors are identical in both columns, so the gain comes from the eigenvalue weighting.

\begin{table}[!t]
\centering
\small
\caption{Cognitive composite from the top-$k$ eigenvectors of each subject's FC, with raw eigenvalues and with eigenvalues raised to $\alpha^{*}$. Mean $\pm$ std over folds (Appendix~\ref{app:mechanism}).}
\label{tab:matched}
\begin{tabular}{lcc}
\toprule
modes & raw $\lambda$ ($\alpha{=}1$) & $\lambda^{\alpha^{*}}$ \\
\midrule
top-1 & $0.321 \pm 0.110$ & $0.321 \pm 0.110$ \\
top-20 & $0.544 \pm 0.079$ & $0.610 \pm 0.079$ \\
top-40 & $0.542 \pm 0.079$ & $0.606 \pm 0.082$ \\
all 400 & $0.542 \pm 0.079$ & $0.624 \pm 0.077$ \\
\bottomrule
\end{tabular}
\end{table}

\subsection{The advantage holds at every data budget tested}
\label{sec:efficiency}

\begin{figure}[!t]
\centering
\includegraphics[width=0.8\linewidth]{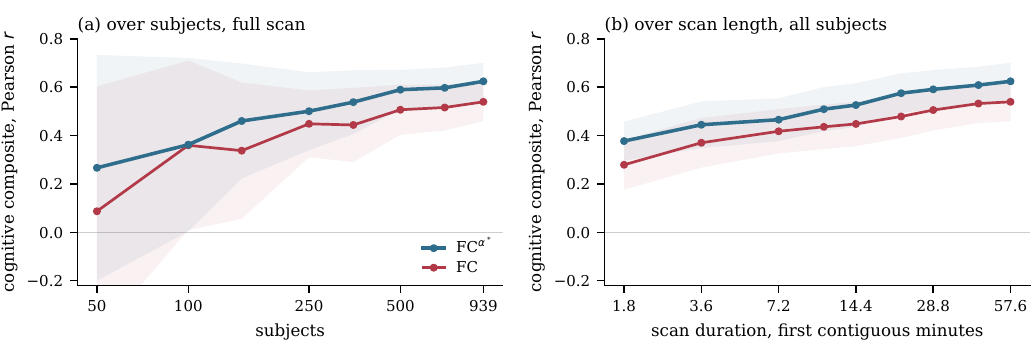}
\caption{Learning curves on HCP-YA Schaefer-400 for $\mathrm{FC}^{\alpha^{*}}$ and raw FC. \textbf{(a)} Against the number of training subjects, at the full scan. \textbf{(b)} Against the duration of the first contiguous minutes of each scan, with all $939$ subjects that have complete runs. Bands are the standard deviation across folds; protocol details in Appendix~\ref{app:protocol}.}
\label{fig:curves}
\end{figure}

We repeat the comparison with fewer training subjects and with shorter scans, plotting the full learning curves (Figure~\ref{fig:curves}). We found that $\mathrm{FC}^{\alpha^{*}}$ is ahead of the standard FC at every number of subjects and at every scan duration.

\subsection{Against brain foundation models}
\label{sec:bench}

\begin{figure}[!t]
\centering
\includegraphics[width=\linewidth]{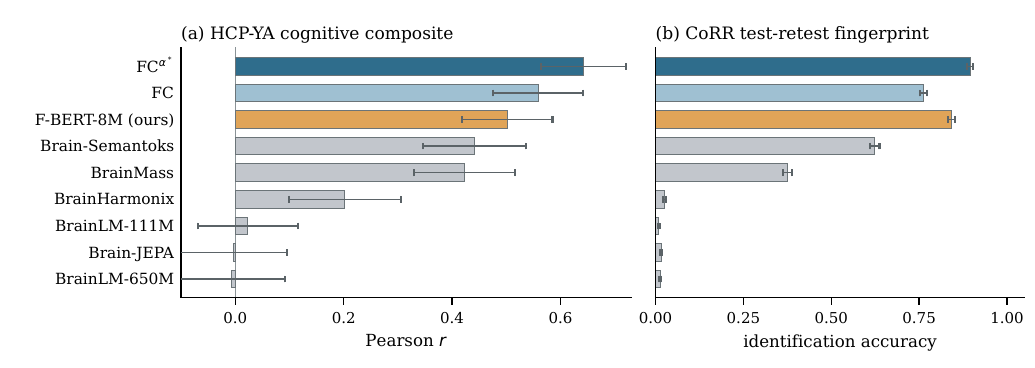}
\caption{Two targets of Table~\ref{tab:bench}: \textbf{(a)} HCP-YA cognitive composite and \textbf{(b)} CoRR identification accuracy, where chance is $1/391$. Error bars are the std over folds in (a) and a subject bootstrap in (b). Blue is FC-derived, orange is ours, grey is published.}
\label{fig:bench}
\end{figure}

Table~\ref{tab:bench} and Figure~\ref{fig:bench} score every published model we could obtain and run, each at the parcellation and window it was released for, under one linear-kernel probe on identical folds; ABIDE-I and the AOMIC-ID1000 replication are in Appendix~\ref{app:bench}.

\begin{table}[!t]
\centering
\small
\caption{Frozen linear-kernel probe (Section~\ref{sec:regression}): HCP-YA cognitive composite (Pearson $r$), sex and age (AUC), and CoRR test--retest fingerprint accuracy, mean $\pm$ std over folds. Within each group the best value per column is bold and, among the foundation models, the second best is underlined; $^{*}$ marks a paired difference with $p<0.05$ (Table~\ref{tab:paired}), against FC at the same scan length for $\mathrm{FC}^{\alpha^{*}}$ and against the best published model of the group for our encoders. \emph{params}: encoder weights, or the probe's ridge weights for the FC rows. Parcellations and windows are those each model was released for (Appendix~\ref{app:bench}); $^{\S}$Brain-Semantoks covers the recording as $16$ evenly spaced $200$\,s crops whose embeddings are averaged to match the protocol of its released probe.}

\label{tab:bench}
\setlength{\tabcolsep}{2.2pt}
\begin{tabular}{@{}lrlrcccc@{}}
\toprule
& & \multicolumn{2}{c}{input} & \multicolumn{3}{c}{HCP-YA} & CoRR \\
\cmidrule(lr){3-4}\cmidrule(lr){5-7}\cmidrule(lr){8-8}
encoder & params & parcels & scan & composite & sex & age & fingerprint \\
\midrule
\multicolumn{8}{l}{\emph{reads the full scan}} \\
$\mathrm{FC}^{\alpha^{*}}$ (ours) & 79.8k & Sch-400 & 57.6m & $\mathbf{0.642 \pm .078}$$^{*}$ & $\mathbf{0.99 \pm .00}$$^{*}$ & $\mathbf{0.63 \pm .04}$ & $\mathbf{0.895 \pm .009}$$^{*}$ \\
FC & 79.8k & Sch-400 & 57.6m & $0.559 \pm .083$ & $0.97 \pm .01$ & $0.62 \pm .03$ & $0.762 \pm .010$ \\
\cmidrule(lr){1-8}
F-BERT-8M (ours) & 8.1M & Sch-450 & 57.6m & $\mathbf{0.502 \pm .083}$$^{*}$ & $\mathbf{0.96 \pm .02}$ & $\mathbf{0.61 \pm .03}$ & $\mathbf{0.841 \pm .009}$$^{*}$ \\
F-BERT-1M (ours) & 0.85M & Sch-450 & 57.6m & $\underline{0.488 \pm .090}$$^{*}$ & $0.94 \pm .02$ & $\underline{0.61 \pm .04}$ & $\underline{0.670 \pm .012}$$^{*}$ \\
Brain-Semantoks & 63M & Sch-457 & 57.6m$^{\S}$ & $0.441 \pm .095$ & $\underline{0.96 \pm .02}$ & $0.59 \pm .04$ & $0.624 \pm .013$ \\
BrainMass & 10.2M & Sch-100 & 57.6m & $0.423 \pm .094$ & $0.93 \pm .02$ & $0.57 \pm .04$ & $0.375 \pm .012$ \\
\midrule
\multicolumn{8}{l}{\emph{reads a window of the scan}} \\
$\mathrm{FC}^{\alpha^{*}}$ (ours) & 101k & Sch-450 & 2.4m & $\mathbf{0.460 \pm .101}$$^{*}$ & $\mathbf{0.91 \pm .03}$$^{*}$ & $\mathbf{0.55 \pm .04}$$^{*}$ & $\mathbf{0.863 \pm .009}$$^{*}$ \\
FC & 101k & Sch-450 & 2.4m & $0.359 \pm .103$ & $0.85 \pm .04$ & $0.52 \pm .04$ & $0.569 \pm .012$ \\
\cmidrule(lr){1-8}
F-BERT-1M (ours) & 0.85M & Sch-450 & 2.4m & $\mathbf{0.418 \pm .100}$$^{*}$ & $0.86 \pm .04$ & $\underline{0.57 \pm .04}$ & $0.513 \pm .013$ \\
F-BERT-8M (ours) & 8.1M & Sch-450 & 2.4m & $\underline{0.412 \pm .100}$ & $\underline{0.87 \pm .04}$ & $0.56 \pm .04$ & $\mathbf{0.705 \pm .012}$$^{*}$ \\
Brain-Semantoks & 63M & Sch-457 & 2.7m & $0.368 \pm .103$ & $\mathbf{0.91 \pm .03}$ & $\mathbf{0.57 \pm .04}$ & $\underline{0.527 \pm .013}$ \\
BrainHarmonix & 88M & Sch-400 & 6.4m & $0.202 \pm .103$ & $0.59 \pm .07$ & $0.56 \pm .04$ & $0.026 \pm .004$ \\
BrainLM-111M & 113M & A424 & 2.4m & $0.023 \pm .093$ & $0.64 \pm .05$ & $0.50 \pm .04$ & $0.009 \pm .002$ \\
Brain-JEPA & 86M & Sch-450 & 1.9m & $-0.003 \pm .099$ & $0.54 \pm .06$ & $0.47 \pm .04$ & $0.015 \pm .003$ \\
BrainLM-650M & 658M & A424 & 2.4m & $-0.007 \pm .098$ & $0.63 \pm .05$ & $0.52 \pm .04$ & $0.013 \pm .003$ \\
\bottomrule
\end{tabular}
\end{table}

Among the rows that read the full scan, $\mathrm{FC}^{\alpha^{*}}$ is first on every dataset and task. Among the published models the four that only read a short window are far below them on the composite regardless of size. Our F-BERT reads the same kind of input as those four: given $2.4$\,min of scan it matches or outperforms the best models on the cognitive composite, where it significantly outperforms raw FC from the same $2.4$\,min. Reading the whole recording our encoder models score above every published model, on the cognitive composite and fingerprinting significantly so (Table~\ref{tab:paired}). On AOMIC-ID1000 we observe the same gains; on ABIDE-I Brain-Semantoks is ahead of both, and on ADHD-200 every representation is near chance (Table~\ref{tab:adhd}).

\textbf{Fingerprinting.} As this transform benefits from recovering variance that distinguishes individual traits, we see far larger gains in fingerprinting and in the cognitive composite than in traits shared by many subjects like sex or age. $\mathrm{FC}^{\alpha^{*}}$ raises identification from $0.762$ to $0.895$ while sex and age move by at most $0.02$, and the trained encoder gains $+0.217$ over the best published model in identification against $+0.007$ and $+0.019$ on sex and age, while many BFMs remain near chance performance at this task (Table~\ref{tab:paired}).

\begin{table}[!htb]
\centering
\caption{Gain of F-BERT-1M over raw FC on the cognitive composite by scan length (rows) and number of subjects (columns), mean $\pm$ std of the paired per-fold difference under the probe of Table~\ref{tab:bench}; $^{*}$ means one-sided $p_{\mathrm{NB}}<0.05$ in favour of the encoder. Green is a gain, red a loss.}
\label{tab:grid}
\setlength{\tabcolsep}{1.6pt}
\begin{tabular}{@{}lcccccc@{}}
\toprule
& \multicolumn{6}{c}{subjects} \\
\cmidrule(lr){2-7}
min & $100$ & $150$ & $250$ & $350$ & $700$ & all \\
\midrule
1.2 & \cellcolor{green!39}$+0.17\pm0.44$ & \cellcolor{green!33}$+0.15\pm0.35$ & \cellcolor{green!23}$+0.10\pm0.25$ & \cellcolor{green!27}$+0.12\pm0.23$ & \cellcolor{green!20}$+0.09\pm0.14$$^{*}$ & \cellcolor{green!19}$+0.09\pm0.11$$^{*}$ \\
2.4 & \cellcolor{green!45}$+0.21\pm0.39$ & \cellcolor{green!40}$+0.18\pm0.34$ & \cellcolor{green!29}$+0.13\pm0.20$$^{*}$ & \cellcolor{green!34}$+0.15\pm0.20$$^{*}$ & \cellcolor{green!18}$+0.08\pm0.11$$^{*}$ & \cellcolor{green!12}$+0.06\pm0.09$$^{*}$ \\
4.8 & \cellcolor{green!45}$+0.22\pm0.37$$^{*}$ & \cellcolor{green!42}$+0.19\pm0.32$$^{*}$ & \cellcolor{green!26}$+0.12\pm0.22$ & \cellcolor{green!19}$+0.09\pm0.17$ & \cellcolor{green!7}$+0.03\pm0.09$ & \cellcolor{green!5}$+0.03\pm0.08$ \\
9.6 & \cellcolor{green!45}$+0.22\pm0.38$$^{*}$ & \cellcolor{green!33}$+0.15\pm0.29$ & \cellcolor{green!19}$+0.09\pm0.19$ & \cellcolor{green!12}$+0.06\pm0.16$ & $+0.00\pm0.08$ & $-0.00\pm0.07$ \\
19.2 & \cellcolor{green!31}$+0.14\pm0.40$ & \cellcolor{green!19}$+0.09\pm0.25$ & \cellcolor{green!13}$+0.06\pm0.17$ & $+0.02\pm0.14$ & $-0.01\pm0.08$ & $-0.02\pm0.06$ \\
38.4 & \cellcolor{green!23}$+0.11\pm0.42$ & \cellcolor{green!10}$+0.05\pm0.28$ & $+0.02\pm0.16$ & \cellcolor{red!6}$-0.03\pm0.13$ & \cellcolor{red!11}$-0.05\pm0.07$ & \cellcolor{red!13}$-0.06\pm0.06$ \\
57.6 & \cellcolor{green!7}$+0.03\pm0.32$ & $+0.01\pm0.24$ & $-0.01\pm0.18$ & $-0.01\pm0.15$ & \cellcolor{red!12}$-0.06\pm0.08$ & \cellcolor{red!15}$-0.07\pm0.06$ \\
\bottomrule
\end{tabular}
\end{table}

\subsection{Distilling the transform into a timeseries model}
\label{sec:distill-results}

The F-BERT rows of Table~\ref{tab:bench} test the pretraining loss of Section~\ref{sec:distill}. The score rises with the duration of the scan and with scale (Appendix~\ref{app:pretrain}). Read over the full recording the encoders are at par with raw FC, but neither reaches its teacher $\mathrm{FC}^{\alpha^{*}}$ (Table~\ref{tab:paired}). On short scans and with fewer subjects the encoders are above KRR on FC (Tables~\ref{tab:grid} and \ref{tab:gridfp}, Figure~\ref{fig:duration}), which is the most common case for studies on individual traits and is precisely where a pretrained model is needed.

The objective transfers part of the transform's structure, enough to pass every published BFM on the composite and to beat raw FC on short scans with fewer subjects. Given that the loss aligns global pairwise distances, we hypothesize a bigger batch should close the gap even further.

\section{Conclusions}

Flattening the eigenvalues of each subject's FC matrix improves prediction and fingerprinting, and the flattened connectome is a usable pretraining target. The transform is one eigendecomposition per subject and can replace the FC matrix in any existing pipeline. It could strengthen the effects found in studies that relate FC to behaviour, diagnosis or identity.

Scanning time and sample size are the main costs of an fMRI study. $\mathrm{FC}^{\alpha^{*}}$ and the encoders stay ahead of FC with fewer subjects and shorter scans (Figure~\ref{fig:curves}, Tables~\ref{tab:bench} and \ref{tab:grid}). This matters most for clinical cohorts, where long scans and large samples are hard to obtain.

Neither the transform nor the released encoders need a GPU (Appendix~\ref{app:bench}). The encoders were pretrained on $162$ openly available datasets from many sites and countries, not on a single national biobank, and we release that data with the models. This makes them a starting point for laboratories that study populations under-represented in the large biobanks.

\section{Limitations}
\label{sec:limitations}

The exponent was selected once, on HCP-YA at Schaefer-400 and on the folds reported for that cell, which is therefore optimistic, and the transform matches or improves prediction on every other dataset, parcellation and target we tested with that same value (Appendix~\ref{app:table1full}, Table~\ref{tab:cohorts}). This does not make $\alpha^{*}$ the optimum for fMRI in general: the best exponent may depend on the dataset, the task and the cohort. Finding this per-cohort optimum $\alpha$ could improve the method further. The pretraining corpus is $162$ publicly available datasets from many sites, countries and protocols, resting state and task, that were never standardised; running them through one preprocessing pipeline means that each dataset is probably preprocessed suboptimally. The reported checkpoints on the encoders were selected by early stopping on the downstream probe rather than trained for a fixed number of steps, so the pretraining loss may start to hurt downstream performance beyond some point. Table~\ref{tab:bench} scores each published model at the parcellation and window length it was released for, so its rows differ in input as well as in model. The transform and the encoders are meant to be used as they are, on small cohorts, on short scans and without a GPU, so every model is compared frozen. Fine-tuning is outside this scope and could change the order of Table~\ref{tab:bench}.

\subsection*{Ethics statement}

This work is a secondary analysis of openly distributed human neuroimaging cohorts. AOMIC-ID1000 \citep{snoek2021amsterdam} is released under CC0 on OpenNeuro; HCP S1200 \citep{VANESSEN201362} is governed by the WU-Minn HCP Open Access Data Use Terms, under which we registered and accepted the no-re-identification clause; family structure, which is HCP restricted data, was used only to keep relatives in the same fold and is not released; ABIDE-I, ADHD-200 and CoRR are distributed de-identified. We performed no new data collection and made no attempt to re-identify participants. Section~\ref{sec:bench} quantifies how identifiable individuals are from their connectivity, and we note that the transform we propose is a strong fingerprint. The released pretraining data are parcellated timeseries and recording identifiers only, derived from datasets already public on OpenNeuro under their own licences (listed in the repository); no images are redistributed, and HCP-YA, ABIDE-I, ADHD-200 and CoRR are not released. All cohorts over-represent Western, healthy, well-educated participants, and the methods here inherit those biases and require re-validation before any clinical use.

\subsection*{Reproducibility statement}

The transform is specified completely in Section~\ref{sec:transform} and has no fitted parameters: one eigendecomposition per subject and one exponent, $\alpha^{*}=0.35$, whose selection is described in Section~\ref{sec:alpha} and Appendix~\ref{app:alpha-choice}. The evaluation protocol, fold seeds, penalty grids and leakage controls are in Section~\ref{sec:eval} and Appendix~\ref{app:protocol}; every comparison between two representations is made on shared folds, and the corrected paired tests are given per cell. The encoder architecture, objective and teacher are in Section~\ref{sec:distill}; the optimiser, learning rates, schedule, batch sizes, window length, checkpoint frequency and the checkpoint-selection rule with its measured optimism are in Appendix~\ref{app:pretrain}. The repository (\url{https://github.com/GioMarraffini/fc-power}) contains the code for every table and figure, one script per table, together with the per-fold results each one renders, so every number in the paper can be checked without the imaging data and recomputed with it; the pretraining loop and its two configurations; the two released encoders at $1$M and $8$M parameters (\url{https://huggingface.co/Marraffini-Giovanni/fbert-1m}, \url{https://huggingface.co/Marraffini-Giovanni/fbert-8m}); and the pretraining data as used (\url{https://huggingface.co/datasets/Marraffini-Giovanni/fbert-corpus}; $162$ open datasets, $9{,}578$ subjects, $27{,}835$ recordings, preprocessed, parcellated and windowed, with the list of recordings), so that the pretraining can be repeated without access to any restricted cohort. The published models we compare with are typically pretrained on cohorts under restricted access, such as the UK Biobank, and released at a single size.

\subsection*{AI use statement}

In this work, we used generative AI tools to help develop theoretical models and conceptual frameworks, formulate mathematical claims, propose and refine hypotheses, design and provide feedback on research methodology and experiments, implement methods, clean and reformat datasets, and interpret results. We have not used generative AI tools to generate synthetic data sets or to discover research topics or identify gaps. Proving mathematical claims and writing proofs, translation, qualitative and thematic data analysis, questions for surveys or interviews and transcription of recordings are not applicable to this work. Additionally, we used generative AI tools to create and modify scientific figures, suggest experimental parameters, create and edit software code, create artifacts, draft parts of this paper, summarise and analyse existing literature, brainstorm, search for information, edit this paper to improve readability, identify relevant literature, format references, suggest a structure for this paper and propose its title and keywords. We have reviewed all AI-assisted work: AI-assisted code was reviewed and tested by the authors, every number in the paper was checked by the authors against the stored experiment outputs, and AI-assisted text and cited literature were checked by the authors. We take responsibility for the final content of this work, including text, claims or artifacts produced with the aid of generative AI.

\subsection*{Acknowledgments}

Data were provided in part by the Human Connectome Project, WU-Minn Consortium (Principal Investigators: David Van Essen and Kamil Ugurbil; 1U54MH091657) funded by the 16 NIH Institutes and Centers that support the NIH Blueprint for Neuroscience Research; and by the McDonnell Center for Systems Neuroscience at Washington University. Experiments presented in this paper were carried out using clusters provided by Inria. This work was granted access to the HPC resources of IDRIS under the allocation 2026-AD011017736 made by GENCI.

\bibliographystyle{iclr2027_conference}
\bibliography{references}

\appendix

\section{Data and preprocessing}
\label{app:data}

\textbf{HCP-YA} \citep{VANESSEN201362}: $955$ adults, resting state, four runs concatenated to $57.6$\,min of scan ($4800$ timepoints at $\mathrm{TR}{=}0.72$\,s); $939$ subjects have four complete runs and are used where scan length must be uniform. S1200 release: minimally preprocessed, ICA-FIX-denoised resting-state runs, parcellated in CIFTI grayordinate space with the Schaefer atlas \citep{schaefer2018local} at $100$, $200$, $300$ and $400$ parcels, and additionally at Schaefer-400 plus the $50$-region Tian-S3 subcortical atlas (Sch-450) and NextBrain-528. Each parcel timeseries is the mean over its vertices, band-passed to $0.01$--$0.08$\,Hz and z-scored per region. $16$ of the $955$ subjects have one incomplete run; the learning curves of Section~\ref{sec:efficiency} use the $939$ with four complete runs, and Appendix~\ref{app:cohort939} gives Table~\ref{tab:main} on that subset. \textbf{AOMIC-ID1000}: fMRIPrep-preprocessed \citep{esteban2019fmriprep}, same parcellations and per-region standardisation; $877$ subjects have the cognitive target and $874$ of them also have every encoder's input for Table~\ref{tab:cohorts}. The Intelligence Structure Test subscales are verbal, numerical and figural reasoning and memory. \textbf{ABIDE-I} \citep{dimartino2014abide}: $1035$ subjects, resting state, $17$ sites, autism diagnosis; $158$ subjects from $4$ sites have every model's input; C-PAC \texttt{filt\_global} derivative; the all-model intersection used in Table~\ref{tab:cohorts} has $n{=}158$. \textbf{ADHD-200} \citep{adhd2012consortium}: $298$ scans from two sites, ADHD diagnosis. \textbf{CoRR} \citep{zuo2014corr}: $391$ subjects with two resting-state sessions, $782$ scans, for test--retest identification. \textbf{Pretraining corpus}: $162$ OpenNeuro datasets ($9{,}578$ subjects, $27{,}835$ recordings), resting state and task, downloaded and preprocessed with fMRIPrep \citep{esteban2019fmriprep} with default settings, then parcellated at Schaefer-400 plus the $50$-region Tian-S3 subcortical atlas ($450$ regions), band-passed and z-scored per region as for HCP-YA. The datasets keep their native repetition times, $0.75$ to $3$\,s, so an $80$-timepoint window spans $1$ to $4$\,min. The parcellated windows, the recording identifiers and both trained encoders are released. Pretraining windows are $80$ timepoints, $106{,}397$ in total, a median of four per recording; at each dataset's repetition time they add up to about $4{,}000$ hours. HCP-YA, ABIDE-I, ADHD-200 and CoRR are not in the corpus.

\section{Evaluation protocols}
\label{app:protocol}

\textbf{Correlation-kernel KRR.} Features are the $P(P-1)/2$ upper-triangle entries of the matrix without the diagonal, weighted by $\sqrt{2}$ so that their Euclidean inner product equals the Frobenius inner product of Equation~\ref{eq:kernel} up to the diagonal term (a constant factor that the correlation kernel removes). The kernel between two subjects is the Pearson correlation of their feature vectors. Hyperparameters are selected by nested cross-validation \citep{varoquaux2017assessing}: the outer loop splits the cohort into $10$ folds, grouped so that members of one family fall in the same fold, and holds one fold out as the test set; inside the remaining $9$ folds an inner $5$-fold cross-validation scores each ridge penalty in $\{10^{-3},\dots,10^{3}\}$ on the pooled inner predictions and selects the best; the model is refit on all $9$ training folds with that penalty and scored once on the held-out fold. The outer loop is repeated $20$ times with different fold assignments, giving $200$ test scores per representation. Every choice, the penalty, the PCA that defines the cognitive composite and the alignment of Section~\ref{sec:alpha}, is made inside the training folds; the held-out fold informs nothing. The exponent grid itself is scored on the outer folds, which is why Section~\ref{sec:alpha} corroborates it with a criterion computed inside training folds only.

\textbf{Linear-kernel ridge probe.} Ridge regression on z-scored features with the same fold structure and penalty selection, on the full $n{=}955$ HCP-YA cohort, site-grouped folds on ABIDE-I and ADHD-200, and nearest-neighbour identification on CoRR. The linear kernel keeps each embedding's norm, which the correlation kernel discards. Raw FC scores $0.543$ under the correlation kernel and $0.559$ under this one ($0.550$ as a mean over folds rather than pooled, Table~\ref{tab:paired}), so values are compared within one protocol only.

\section{The scaled eigenvectors before and after the transform}
\label{app:sphere}

\begin{figure}[!htb]
\centering
\includegraphics[width=\linewidth]{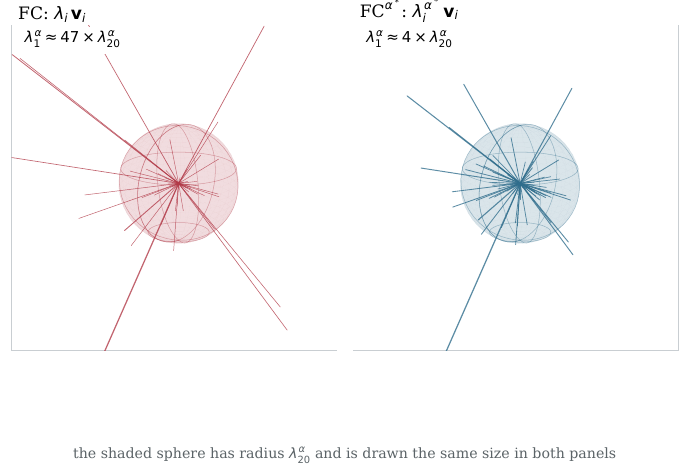}
\caption{The scaled eigenvectors $\lambda_i\mathbf{v}_i$ (left) and $\lambda_i^{\alpha^{*}}\mathbf{v}_i$ (right) of the mean HCP-YA Schaefer-400 spectrum in a random three-dimensional projection, both drawn at one scale, set so the shaded sphere of radius $\lambda_{20}^{\alpha}$ is the same size in each. In raw FC a few modes extend far beyond the sphere and the remaining modes are shorter than a small fraction of its radius; after the transform no mode exceeds four sphere radii.}
\end{figure}

\section{The exponent}
\label{app:alpha}

\subsection{The transform as a geodesic}
\label{app:geodesic}

Symmetric positive definite (SPD) matrices form a curved space, and the two metrics used for connectomes measure distance on it through the matrix logarithm. Under the log-Euclidean metric \citep{arsigny2006logeuclidean} the distance between two SPD matrices is $d(\mathbf{A},\mathbf{B})=\|\log\mathbf{A}-\log\mathbf{B}\|_F$ and the geodesic from $\mathbf{A}$ to $\mathbf{B}$ is $\gamma(\alpha)=\exp\big((1-\alpha)\log\mathbf{A}+\alpha\log\mathbf{B}\big)$. Taking $\mathbf{A}=\mathbf{I}$ and $\mathbf{B}=\mathbf{\Sigma}$ gives $\gamma(\alpha)=\exp(\alpha\log\mathbf{\Sigma})=\mathbf{\Sigma}^{\alpha}$, so the power family is the geodesic from the identity to the raw connectome. Under the affine-invariant metric \citep{pennec2006riemannian} the geodesic from $\mathbf{A}$ to $\mathbf{B}$ is $\mathbf{A}^{1/2}(\mathbf{A}^{-1/2}\mathbf{B}\mathbf{A}^{-1/2})^{\alpha}\mathbf{A}^{1/2}$, which for $\mathbf{A}=\mathbf{I}$ is again $\mathbf{\Sigma}^{\alpha}$; the two metrics agree on this path because it starts at the identity. The parameter is arclength: $d(\mathbf{I},\mathbf{\Sigma}^{\alpha})=\alpha\,\|\log\mathbf{\Sigma}\|_F$ and $d(\mathbf{\Sigma}^{\alpha},\mathbf{\Sigma})=(1-\alpha)\|\log\mathbf{\Sigma}\|_F$ (Figure~\ref{fig:geo}b), so $\alpha^{*}$ is the point $35\%$ of the way from the identity to the raw matrix. In the eigenbasis the geodesic moves each eigenvalue along a straight line in $\log\lambda$, from $0$ at $\alpha=0$ to $\log\lambda_i$ at $\alpha=1$ (Figure~\ref{fig:geo}a); eigenvalues above $1$ descend and eigenvalues below $1$ rise, at a rate proportional to $|\log\lambda_i|$, which is why the extremes of the spectrum move most.

The tangent vector of this geodesic at $\alpha=0$ is $\frac{d}{d\alpha}\mathbf{\Sigma}^{\alpha}\big|_{0}=\log\mathbf{\Sigma}$, which is the feature of the tangent-space parameterisation of Section~\ref{sec:related} with the identity as reference. That method therefore represents each subject by the direction of departure from the identity, whereas $\mathrm{FC}^{\alpha^{*}}$ represents it by a point part of the way along the same path. Since $(\mathbf{\Sigma}^{\alpha}-\mathbf{I})/\alpha\to\log\mathbf{\Sigma}$, the log map is the $\alpha\to0$ endpoint of the family, and Appendix~\ref{app:screen} places it below the interior optimum. With a group reference $\mathbf{\Sigma}_{\mathrm{ref}}\neq\mathbf{I}$ the tangent map also whitens by $\mathbf{\Sigma}_{\mathrm{ref}}^{-1/2}$, which rotates each subject's eigenbasis; the geodesic from the identity does not.

\subsection{The transform as a heat kernel}
\label{app:heat}

Write $\lambda^{\alpha}=e^{\alpha\log\lambda}$. With $\mathbf{L}=-\log\mathbf{\Sigma}=\mathbf{V}\,\mathrm{diag}(-\log\lambda_i)\mathbf{V}^{\top}$, the transformed matrix is $\mathbf{\Sigma}^{\alpha}=e^{-\alpha\mathbf{L}}$, and $\mathbf{H}(t)=e^{-t\mathbf{L}}$ is the solution of the heat equation $\partial_t\mathbf{H}=-\mathbf{L}\mathbf{H}$ with $\mathbf{H}(0)=\mathbf{I}$. For a correlation matrix $\mathbf{L}$ is indefinite, since $-\log\lambda_i<0$ for the eigenvalues above $1$. Dividing by $\lambda_1$ fixes this: $\tilde{\mathbf{L}}=-\log(\mathbf{\Sigma}/\lambda_1)$ has eigenvalues $\tilde\ell_i=\log(\lambda_1/\lambda_i)\ge0$ with $\tilde\ell_1=0$, and $\tilde{\mathbf{H}}(t)=e^{-t\tilde{\mathbf{L}}}=(\mathbf{\Sigma}/\lambda_1)^{t}=\lambda_1^{-t}\mathbf{\Sigma}^{t}$ is a proper heat semigroup: the identity at $t=0$, the raw connectome up to scale at $t=1$, and the rank-one projector $\mathbf{v}_1\mathbf{v}_1^{\top}$ as $t\to\infty$. The per-subject factor $\lambda_1^{-t}$ scales every feature of that subject by one constant, which the Pearson correlation kernel of the regression removes, so under the correlation-kernel KRR $\mathbf{\Sigma}^{\alpha}$ and $\tilde{\mathbf{H}}(\alpha)$ give the same kernel; under the z-scored linear probe of Table~\ref{tab:bench} they do not, and the reading is exact only for the first regression.

\begin{figure}[!htb]
\centering
\includegraphics[width=\linewidth]{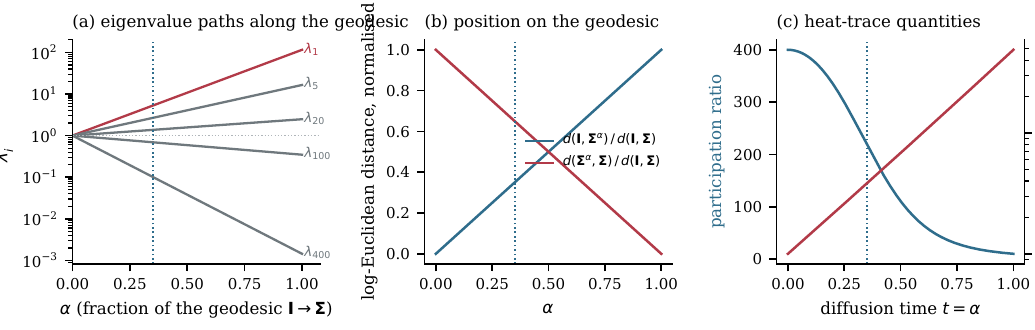}
\caption{The geodesic and heat-kernel interpretations on the $955$ HCP-YA Schaefer-400 spectra. \textbf{(a)} Eigenvalues of the mean spectrum along the geodesic $\mathbf{\Sigma}^{\alpha}$, straight lines in $\log\lambda$ from $1$ at $\alpha=0$ to $\lambda_i$ at $\alpha=1$. \textbf{(b)} Log-Euclidean distance of $\mathbf{\Sigma}^{\alpha}$ from the identity and from the raw matrix, as fractions of $d(\mathbf{I},\mathbf{\Sigma})$; both are linear in $\alpha$. \textbf{(c)} Participation ratio $Z(t)^2/Z(2t)$ and first-to-twentieth eigenvalue ratio $(\lambda_1/\lambda_{20})^{t}$ against diffusion time, means over subjects. The dotted line is $\alpha^{*}$ in every panel.}
\label{fig:geo}
\end{figure}

In this reading $\alpha$ is a diffusion time. The eigenvalues $\tilde\ell_i$ are decay rates: the first mode does not decay, the tenth decays at rate $\log(\lambda_1/\lambda_{10})$, and the mode with the smallest eigenvalue decays fastest. At $t=1$ the fast modes have decayed by factors of up to $\lambda_1/\lambda_{400}\approx10^{5}$, so almost all of the trace sits on the ten slowest modes; at $t=0$ nothing has decayed and every mode has weight one; $\alpha^{*}$ is the snapshot at which the slow modes lead but the rest are still present. The quantities of Section~\ref{sec:transform} are functions of the heat trace $Z(t)=\mathrm{tr}\,\tilde{\mathbf{H}}(t)=\sum_i(\lambda_i/\lambda_1)^{t}$: the participation ratio is $\mathrm{PR}(t)=Z(t)^2/Z(2t)$ and the dominance of the first mode over the twentieth is $e^{t\tilde\ell_{20}}=(\lambda_1/\lambda_{20})^{t}$. Figure~\ref{fig:geo}c plots both against $t$ over the $955$ subjects; at $t=\alpha^{*}$ they give $218.9$ and $3.83$, the first quoted in Section~\ref{sec:transform}, and at $t=1$ they return to $10.1$ and $48.9$. The regression kernel of Equation~\ref{eq:kernel} becomes the overlap of two subjects' heat kernels at the same time, $\mathrm{tr}\big(\tilde{\mathbf{H}}_a(t)\tilde{\mathbf{H}}_b(t)\big)=\sum_{i,j}e^{-t(\tilde\ell^{a}_i+\tilde\ell^{b}_j)}(\mathbf{v}_i^{\top}\mathbf{u}_j)^2$, in which the weight on the pair of modes $(i,j)$ decays at the sum of their two rates.

\subsection{The transform as a spectral filter}
\label{app:filter}

In signal processing on graphs \citep{shuman2013emerging} a filter is a function $h$ applied to the eigenvalues of an operator while its eigenvectors are kept: $h(\mathbf{\Sigma})=\mathbf{V}\,\mathrm{diag}(h(\lambda_i))\mathbf{V}^{\top}$. The power transform is the filter $h(\lambda)=\lambda^{\alpha}$ on the eigenmodes of the connectome itself, and the heat kernel of Appendix~\ref{app:heat} is the same filter written on the generator: $e^{-t\tilde\ell_i}$ with $\tilde\ell_i=\log(\lambda_1/\lambda_i)$. On a graph Laplacian the heat kernel $e^{-t\ell}$ is the standard low-pass filter, since it attenuates the modes with large $\ell$, the graph frequencies; here $\tilde\ell_i$ orders the modes by decreasing variance, so ``high frequency'' means low-variance eigenmode, and the frequency axis is the eigenvalue rank of $\mathbf{\Sigma}$, not the temporal frequency of the BOLD signal, which the band-pass of Appendix~\ref{app:data} fixes separately. In this sense every $\mathbf{\Sigma}^{t}$ with $t>0$ is a low-pass filter on the connectome's modes, and $t$ sets how steeply the attenuation falls with rank: at $t=1$ the filter is $\lambda_i/\lambda_1$, which passes the first mode at unity and the four-hundredth at $10^{-5}$; at $t=0$ it is all-pass; at $\alpha^{*}$ the same span is $(10^{-5})^{0.35}\approx2\times10^{-2}$.

Relative to raw FC, which is the representation the regression is normally given, the transform is therefore not a low-pass but its partial inverse. Writing $\mathbf{\Sigma}^{\alpha}=g(\mathbf{\Sigma})$ with $g(\lambda)=\lambda^{\alpha-1}\cdot\lambda$, the factor $g(\lambda)/\lambda=\lambda^{\alpha-1}$ is a monotone decreasing gain that amplifies the low-variance modes relative to the high-variance ones, a spectral equaliser that compresses the dynamic range of the modes from $10^{5}$ to about $50$ without changing their order; full equalisation, $\alpha=0$, is the whitening filter $\lambda^{-1}$, which maps every subject's matrix to the identity. The filter is monotone, so it never reorders modes, and Appendix~\ref{app:mechanism} shows that non-monotone filters, which would boost a band of modes above its neighbours, do not improve on it.

\subsection{Computing the alignment optimum}
\label{app:alpha-choice}

\begin{figure}[!htb]
\centering
\includegraphics[width=0.78\linewidth]{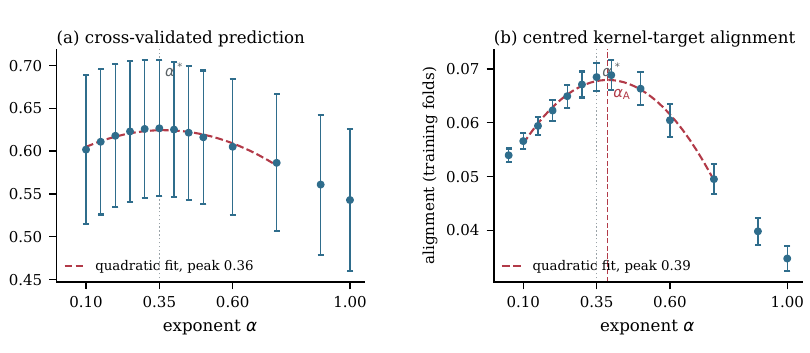}
\caption{Two of the criteria of Section~\ref{sec:alpha} on HCP-YA at Schaefer-400, from the same eigendecompositions and folds. \textbf{(a)} Cognitive composite under the correlation-kernel KRR, mean over the $200$ folds with the standard deviation across folds as the bar. \textbf{(b)} Centred kernel-target alignment with the cognitive composite inside training folds, mean and standard deviation over the $200$ folds; the dashed vertical line is its continuous optimum $\alpha_{\mathrm{A}}=0.387$. In each panel the dashed curve is a variance-weighted quadratic fit over $\alpha\in[0.10,0.75]$ with its peak marked, and the dotted vertical line is $\alpha^{*}$.}
\label{fig:alpha}
\end{figure}

Every exponent shares the eigenvectors, so the whole search runs on one eigendecomposition per subject: $\mathbf{\Sigma}_s=\mathbf{V}_s\mathbf{D}_s\mathbf{V}_s^{\top}$ is computed once ($O(P^3)$ per subject, $955$ subjects at $P=400$), and for any $\alpha$ the feature vector $\mathrm{vec}(\mathbf{V}_s\mathbf{D}_s^{\alpha}\mathbf{V}_s^{\top})$ and the kernel $\mathbf{K}_{\alpha}$ follow by re-weighting. The alignment $\mathrm{A}(\alpha)$ of Equation~\ref{eq:kta} is then a smooth scalar function of one variable, and its derivative is available in closed form through $\partial_{\alpha}\mathbf{\Sigma}^{\alpha}=\mathbf{V}\,\mathrm{diag}(\lambda_i^{\alpha}\log\lambda_i)\mathbf{V}^{\top}$ propagated through the per-subject centring and normalisation of the correlation kernel and through the centring $\mathbf{H}$. We maximise $\mathrm{A}$ with Brent's method \citep{brent1973algorithms}, a bracketing search that combines golden-section steps with parabolic interpolation and needs no derivative; on $(0,1]$ it converges to a tolerance of $10^{-4}$ in $8$ evaluations, fewer than the $13$-point grid of Figure~\ref{fig:alpha}b, and the analytic derivative, evaluated at the optimum as a check, agrees with a central finite difference to a relative error of $1.2\times10^{-7}$. The optimum $\alpha_{\mathrm{A}}=0.387$ is computed on training folds only and averaged over the $200$ folds of the correlation-kernel protocol; the whole computation, eigendecompositions included, takes about half an hour on $16$ CPU cores.

\section{The transform screen}
\label{app:screen}

Table~\ref{tab:screen-short} compares $\mathrm{FC}^{\alpha^{*}}$ with raw FC and with three standard transformations of the FC matrix on HCP-YA Schaefer-400 ($n{=}955$), under the correlation-kernel KRR of Section~\ref{sec:eval} with $10$ repetitions of $10$ family-aware folds. The log-Euclidean map \citep{arsigny2006logeuclidean} takes the matrix logarithm of each subject's matrix and is the $\alpha\to0$ limit of the power family. The tangent-space parameterisation \citep{varoquaux2010detection} takes the matrix logarithm after whitening each subject's matrix by a group reference, here the harmonic mean of the cohort. Partial correlation is the correlation between two regions given all the others, computed from the inverse of the Ledoit--Wolf shrunk covariance \citep{ledoit2004wellconditioned}.

\begin{table}[!htb]
\centering
\small
\caption{$\mathrm{FC}^{\alpha^{*}}$ against raw FC and three standard transformations. HCP-YA Schaefer-400, cognitive composite, correlation-kernel KRR, $10\times10$ family-aware CV, mean $\pm$ std over the $100$ folds. $^{*}$ marks a row significantly below $\mathrm{FC}^{\alpha^{*}}$ on shared folds, $p_{\mathrm{NB}}<0.05$ under the corrected resampled $t$-test \citep{nadeau2003inference}.}
\label{tab:screen-short}
\begin{tabular}{lc}
\toprule
transform & $r$ \\
\midrule
$\mathrm{FC}^{\alpha^{*}}$ & $\mathbf{0.628 \pm 0.077}$ \\
log-Euclidean & $0.586 \pm 0.085^{*}$ \\
tangent space, harmonic reference & $0.578 \pm 0.083^{*}$ \\
raw FC & $0.545 \pm 0.083^{*}$ \\
partial correlation & $0.389 \pm 0.100^{*}$ \\
\bottomrule
\end{tabular}
\end{table}

Before the exponent grid, we compared $\mathrm{FC}^{\alpha}$ against other transforms of the FC matrix on HCP-YA Schaefer-400 ($n{=}955$), under the correlation-kernel KRR of Section~\ref{sec:eval} with $10$ repetitions of $10$ family-aware folds. Group-fit transforms (tangent maps, PCA, ICA, factor analysis) were fit once on the full cohort in this screen; refitting them inside each fold, or on a disjoint external corpus of $8{,}000$ recordings, changed their scores by at most $0.004$ (tangent $0.578$ fit-once against $0.574$ per-fold), so the ranking does not depend on where they are fit. Raw FC scores $0.545$ here against $0.543$ under the $20$-repetition protocol of Table~\ref{tab:main}.

\begin{table}[!htb]
\centering
\small
\caption{Transforms of the HCP-YA Schaefer-400 FC matrix, cognitive composite, correlation-kernel KRR, $10\times10$ family-aware CV, mean $\pm$ std over folds. Power-Euclidean rows and the log-Euclidean map are per-subject; tangent maps use a group reference (harmonic mean or corpus mean); the remaining rows are covariance estimators, edge normalisations and subspace reductions.}
\label{tab:screen}
\footnotesize\setlength{\tabcolsep}{3pt}
\begin{tabular}{lc@{\hskip 6pt}lc}
\toprule
transform & $r$ & transform & $r$ \\
\midrule
$\mathrm{FC}^{0.25}$ & $0.622 \pm 0.079$ & Fisher $z$ of the edges & $0.555 \pm 0.082$ \\
$\mathrm{FC}^{0.5}$ (matrix square root) & $0.619 \pm 0.077$ & degree normalisation & $0.554 \pm 0.084$ \\
$\mathrm{FC}^{0.75}$ & $0.589 \pm 0.080$ & raw FC ($\alpha{=}1$) & $0.545 \pm 0.083$ \\
log-Euclidean ($\alpha\to0$) & $0.586 \pm 0.085$ & PCA, $300$ components & $0.539 \pm 0.086$ \\
tangent, harmonic reference & $0.578 \pm 0.083$ & factor analysis / ICA & $0.510 \pm 0.081$ / $0.503 \pm 0.084$ \\
tangent, corpus reference & $0.578 \pm 0.086$ & Cholesky / log-Cholesky & $0.503 \pm 0.090$ / $0.502 \pm 0.094$ \\
Tucker-denoised FC & $0.573 \pm 0.082$ & random projection & $0.504 \pm 0.086$ \\
tangent, affine-invariant & $0.562 \pm 0.090$ & precision / partial correlation & $0.410 \pm 0.094$ / $0.389 \pm 0.100$ \\
\bottomrule
\end{tabular}
\end{table}

Every transform above $0.58$ re-weights the spectrum of each subject's own matrix, and every subspace reduction, edge normalisation or estimator that does not re-weight the spectrum scores at or below raw FC. Within the power family the score rises from $\alpha=1$ to a maximum between $0.25$ and $0.5$ and falls again toward the log-Euclidean limit, which is the result Figure~\ref{fig:alpha} refines on the finer grid. The fine grid of Figure~\ref{fig:alpha} was run afterwards at $\{0.10,0.15,\dots,0.50\}$ on this cohort and on AOMIC, at four Schaefer resolutions each, under the $20\times10$ protocol of Table~\ref{tab:main}.

\section{Where the gain comes from: controls at matched eigenvectors}
\label{app:mechanism}

All results here are on HCP-YA Schaefer-400, the $939$ subjects with four complete runs, correlation-kernel KRR with family-aware folds (three repetitions of $10$), cognitive composite; raw FC scores $0.542\pm0.079$ and $\mathrm{FC}^{\alpha^{*}}$ $0.624\pm0.077$ under this protocol.

\paragraph{Per-mode and cumulative prediction.} Reconstructing each subject's matrix from a single eigenvector, $\lambda_i^{\alpha^{*}}\mathbf{v}_i\mathbf{v}_i^{\top}$, and predicting from it alone gives $0.321\pm0.110$ for the first mode, $0.261$ for the fourth, $0.175$ for the tenth, $0.111$ for the twentieth and $0.057$ for the fiftieth. Cumulatively, the top $1$, $5$, $20$, $100$ and all $400$ modes give $0.321$, $0.532$, $0.610$, $0.617$ and $0.624$. The signal is distributed over the top $\sim20$ modes, none of which is strong alone.

\paragraph{Same eigenvectors, two weightings.} Table~\ref{tab:matched} scores the same top-$k$ eigenvectors with raw eigenvalues and with $\lambda^{\alpha^{*}}$. With one mode the two are identical, since a scalar cancels in the normalised kernel. With $20$ modes the re-weighting adds $+0.066$, and raw FC saturates at $20$ modes: adding the remaining $380$ changes it by $-0.002$, because under raw weights they receive almost no kernel weight. The eigenvectors are the same in both columns, so the difference is the eigenvalue weighting alone.

\paragraph{Other spectral filters.} Replacing each subject's eigenvalues by a fixed hand-set profile (top $20$ modes from $10$ down to $1$, middle modes $0.01$, tail from $0.5$ to $0.05$; same eigenvectors) scores $0.586\pm0.081$, so compressing the top alone recovers about half of the gain and the subject's own compressed magnitudes recover the rest. Clipping eigenvalues at $5$ scores $0.615\pm0.087$. Of $100$ random monotone and non-monotone filters $g(\lambda)$, none exceeds $0.621$, and every top performer is monotone. Filters optimised directly on the prediction loss inside nested cross-validation, a $30$-parameter binned filter initialised at $\lambda^{\alpha^{*}}$ and a $64\times64$ MLP, score $0.586\pm0.090$ and $0.579\pm0.083$ against $0.624$ on the same folds; the extra capacity fits the inner objective and generalises worse.

\paragraph{The kernel form does not matter; the per-subject basis does.} On $\mathrm{FC}^{\alpha^{*}}$ features, Pearson, cosine, centred and raw dot-product kernels agree to $0.0006$. Replacing the inner product by an RBF kernel on the geodesic distance of each SPD metric is at or below the linear kernel for every metric (power-Euclidean $0.620$, log-Euclidean $0.573$). A subject's own top-$20$ eigenvectors keep $0.610$, whereas a shared basis (population PCA of the $\mathrm{FC}^{\alpha^{*}}$ features) needs about $200$ components to match it and $900$ components give $0.589$: the informative directions differ across subjects, which is why per-subject re-weighting outperforms any shared or learned projection.

\paragraph{Not an identity effect.} Features that identify subjects perfectly but carry no brain structure predict cognition at chance under family-aware folds: a one-hot subject identifier gives $0.000$ and a random fixed per-subject embedding $0.054\pm0.023$. $\mathrm{FC}^{\alpha^{*}}$ raises the split-half reliability of the out-of-fold predictions from $0.597\pm0.010$ (raw FC) to $0.830\pm0.001$; its disattenuated ceiling is $0.655$, and $0.624$ is $95\%$ of it.

\section{Table~\ref{tab:main} at every parcellation}
\label{app:table1full}

\begin{table}[!htb]
\centering
\small
\caption{Table~\ref{tab:main} at every parcellation. Pearson $r$, mean $\pm$ std over the $200$ fold-level values of a $20\times10$ replicated CV. $\Delta$ is the paired gain on shared folds and $p_{\mathrm{NB}}$ its corrected resampled $t$-test \citep{nadeau2003inference}. Sixteen HCP-YA recordings are shorter than the rest; the same table restricted to the $939$ subjects with four complete runs is Appendix~\ref{app:cohort939}.}
\label{tab:mainfull}
\begin{tabular}{lcccc}
\toprule
parcellation & FC ($\alpha{=}1$) & $\mathrm{FC}^{\alpha^{*}}$ & $\Delta$ & $p_{\mathrm{NB}}$ \\
\midrule
\multicolumn{5}{l}{\emph{HCP-YA, resting state, $N{=}955$}} \\
Schaefer-100 & $0.505 \pm 0.088$ & $0.575 \pm 0.081$ & $+0.070$ & $1{\times}10^{-6}$ \\
Schaefer-200 & $0.519 \pm 0.080$ & $0.595 \pm 0.078$ & $+0.076$ & $2{\times}10^{-8}$ \\
Schaefer-300 & $0.527 \pm 0.084$ & $0.618 \pm 0.077$ & $+0.090$ & $2{\times}10^{-10}$ \\
Schaefer-400 & $0.543 \pm 0.083$ & $\mathbf{0.627 \pm 0.080}$ & $+0.084$ & $2{\times}10^{-9}$ \\
Schaefer-400 + Tian-S3 & $0.519 \pm 0.083$ & $0.574 \pm 0.083$ & $+0.055$ & $5{\times}10^{-5}$ \\
NextBrain-528 & $0.404 \pm 0.088$ & $0.461 \pm 0.082$ & $+0.057$ & $7{\times}10^{-4}$ \\
\midrule
\multicolumn{5}{l}{\emph{AOMIC-ID1000, movie watching, $N{=}877$}} \\
Schaefer-100 & $0.310 \pm 0.091$ & $0.409 \pm 0.083$ & $+0.099$ & $2{\times}10^{-7}$ \\
Schaefer-200 & $0.311 \pm 0.088$ & $0.435 \pm 0.082$ & $+0.124$ & $3{\times}10^{-9}$ \\
Schaefer-300 & $0.334 \pm 0.083$ & $0.444 \pm 0.080$ & $+0.109$ & $4{\times}10^{-7}$ \\
Schaefer-400 & $0.348 \pm 0.084$ & $\mathbf{0.432 \pm 0.081}$ & $+0.084$ & $4{\times}10^{-5}$ \\
\bottomrule
\end{tabular}
\end{table}

\section{The complete-scan subset}
\label{app:cohort939}

Sixteen HCP-YA recordings have one incomplete run. Table~\ref{tab:main} keeps them; restricting to the $939$ subjects with four complete runs moves every cell by at most $0.011$.

\begin{table}[!htb]
\centering
\small
\caption{Table~\ref{tab:main} restricted to the $939$ HCP-YA subjects whose four runs are all complete. AOMIC is unchanged and omitted.}
\begin{tabular}{lcccc}
\toprule
parcellation & FC ($\alpha{=}1$) & $\mathrm{FC}^{\alpha^{*}}$ & $\Delta$ & $p_{\mathrm{NB}}$ \\
\midrule
\multicolumn{5}{l}{\emph{HCP-YA, resting state, $N{=}939$}} \\
Schaefer-100 & $0.494 \pm 0.086$ & $0.567 \pm 0.083$ & $+0.073$ & $3{\times}10^{-7}$ \\
Schaefer-200 & $0.509 \pm 0.086$ & $0.588 \pm 0.080$ & $+0.079$ & $2{\times}10^{-7}$ \\
Schaefer-300 & $0.522 \pm 0.088$ & $0.614 \pm 0.079$ & $+0.092$ & $6{\times}10^{-9}$ \\
Schaefer-400 & $0.540 \pm 0.085$ & $\mathbf{0.624 \pm 0.080}$ & $+0.083$ & $3{\times}10^{-8}$ \\
\bottomrule
\end{tabular}
\end{table}

\section{Foundation-model comparison details}
\label{app:bench}

Each published model was run from its released checkpoint with its released feature-extraction code, at the parcellation and window length it was trained for (Table~\ref{tab:bench}, \emph{input} columns): BrainMass on Schaefer-100 FC matrices; BrainLM ($111$M and $650$M) on AAL-424 timeseries in $144$\,s windows; Brain-JEPA on Sch-450 timeseries in $115$\,s windows; BrainHarmonix on the $17$-network ordering of Schaefer-400, $11$ windows of $35$\,s. Window embeddings are averaged per subject before probing. The ABIDE-I cohort is the intersection of subjects for which every model's input could be produced ($n{=}158$). Table~\ref{tab:cohorts} gives AOMIC-ID1000, ABIDE-I and ADHD-200 for every encoder.

Table~\ref{tab:paired} gives the paired tests between the encoders of Table~\ref{tab:bench}.

\paragraph{Inference cost.} Measured on four CPU threads with no GPU, one HCP-YA recording of $57.6$\,min ($4800$ timepoints, Sch-450) read as one sequence: the $1$M model takes $0.46$\,s and under $1$\,GB of memory, the $8$M architecture ($384$d, $8$L) $2.6$\,s and $1.4$\,GB; a $200$-timepoint window takes $30$\,ms.

\begin{table}[!htb]
\centering
\small
\caption{Paired tests behind Table~\ref{tab:bench}, HCP-YA cognitive composite. Table~\ref{tab:bench} reports the correlation of the pooled out-of-fold predictions, averaged over the $20$ repetitions; a paired test needs one value per fold, so it uses the $200$ per-fold correlations of the same runs, whose mean is given next to the pooled value (top). Bottom: for each pair and target, the difference of per-fold means and its corrected resampled $t$-test $p_{\mathrm{NB}}$ \citep{nadeau2003inference}, over $200$ folds for the composite and $100$ ($10\times10$) for sex and age; for the CoRR fingerprint, the difference in identification accuracy over the $782$ shared scans and an exact McNemar test on the discordant scans. Bold: first encoder significantly ahead ($p<0.05$), the rule that places the asterisks of Table~\ref{tab:bench}.}
\label{tab:paired}
\footnotesize\setlength{\tabcolsep}{4pt}
\begin{tabular}{lcc}
\toprule
encoder & pooled $r$ (Table~\ref{tab:bench}) & per-fold $r$ \\
\midrule
$\mathrm{FC}^{\alpha^{*}}$ & $0.642$ & $0.629 \pm 0.078$ \\
FC & $0.559$ & $0.550 \pm 0.083$ \\
F-BERT-8M & $0.502$ & $0.520 \pm 0.083$ \\
F-BERT-1M & $0.488$ & $0.491 \pm 0.090$ \\
Brain-Semantoks & $0.441$ & $0.437 \pm 0.095$ \\
BrainMass & $0.423$ & $0.433 \pm 0.094$ \\
$\mathrm{FC}^{\alpha^{*}}$, 2.4m & $0.460$ & $0.439 \pm 0.101$ \\
FC, 2.4m & $0.359$ & $0.349 \pm 0.103$ \\
F-BERT-1M, 2.4m & $0.418$ & $0.408 \pm 0.100$ \\
F-BERT-8M, 2.4m & $0.412$ & $0.404 \pm 0.100$ \\
Brain-Semantoks, 2.7m & $0.368$ & $0.362 \pm 0.103$ \\
\bottomrule
\end{tabular}
\\[6pt]
\scriptsize\setlength{\tabcolsep}{3pt}
\begin{tabular}{@{}llcccc@{}}
\toprule
& & \multicolumn{3}{c}{HCP-YA} & CoRR \\
\cmidrule(lr){3-5}\cmidrule(lr){6-6}
first & second & composite & sex & age & fingerprint \\
\midrule
$\mathrm{FC}^{\alpha^{*}}$ & FC & \textbf{$+0.079$ ($2{\times}10^{-13}$)} & \textbf{$+0.020$ ($4{\times}10^{-8}$)} & $+0.015$ (0.14) & \textbf{$+0.133$ ($<10^{-15}$)} \\
$\mathrm{FC}^{\alpha^{*}}$ & F-BERT-8M & \textbf{$+0.109$ ($4{\times}10^{-8}$)} & \textbf{$+0.029$ ($2{\times}10^{-7}$)} & $+0.023$ (0.08) & \textbf{$+0.054$ ($7{\times}10^{-6}$)} \\
$\mathrm{FC}^{\alpha^{*}}$ & F-BERT-1M & \textbf{$+0.138$ ($1{\times}10^{-10}$)} & \textbf{$+0.053$ ($2{\times}10^{-10}$)} & \textbf{$+0.028$ (0.03)} & \textbf{$+0.225$ ($<10^{-15}$)} \\
F-BERT-8M & Brain-Semantoks & \textbf{$+0.083$ ($3{\times}10^{-5}$)} & $+0.007$ (0.31) & $+0.019$ (0.11) & \textbf{$+0.217$ ($<10^{-15}$)} \\
F-BERT-8M & BrainMass & \textbf{$+0.087$ ($8{\times}10^{-4}$)} & \textbf{$+0.038$ ($3{\times}10^{-6}$)} & \textbf{$+0.041$ ($5{\times}10^{-3}$)} & \textbf{$+0.467$ ($<10^{-15}$)} \\
F-BERT-1M & Brain-Semantoks & \textbf{$+0.054$ ($2{\times}10^{-3}$)} & $-0.016$ ($7{\times}10^{-3}$) & $+0.014$ (0.18) & \textbf{$+0.046$ ($7{\times}10^{-3}$)} \\
F-BERT-1M & BrainMass & \textbf{$+0.058$ (0.03)} & $+0.015$ (0.11) & \textbf{$+0.036$ (0.01)} & \textbf{$+0.295$ ($<10^{-15}$)} \\
F-BERT-8M & F-BERT-1M & $+0.029$ (0.07) & \textbf{$+0.023$ ($2{\times}10^{-4}$)} & $+0.005$ (0.66) & \textbf{$+0.171$ ($<10^{-15}$)} \\
FC & F-BERT-8M & $+0.031$ (0.13) & $+0.009$ (0.10) & $+0.008$ (0.55) & $-0.079$ ($2{\times}10^{-7}$) \\
FC & F-BERT-1M & \textbf{$+0.059$ ($7{\times}10^{-3}$)} & \textbf{$+0.032$ ($3{\times}10^{-5}$)} & $+0.013$ (0.33) & \textbf{$+0.092$ ($2{\times}10^{-8}$)} \\
$\mathrm{FC}^{\alpha^{*}}$, 2.4m & FC, 2.4m & \textbf{$+0.090$ ($2{\times}10^{-6}$)} & \textbf{$+0.059$ ($6{\times}10^{-11}$)} & \textbf{$+0.032$ ($1{\times}10^{-3}$)} & \textbf{$+0.294$ ($<10^{-15}$)} \\
F-BERT-1M, 2.4m & FC, 2.4m & \textbf{$+0.059$ (0.04)} & $+0.009$ (0.56) & \textbf{$+0.039$ ($1{\times}10^{-2}$)} & $-0.056$ ($8{\times}10^{-3}$) \\
F-BERT-8M, 2.4m & FC, 2.4m & \textbf{$+0.055$ (0.04)} & $+0.024$ (0.11) & \textbf{$+0.028$ (0.05)} & \textbf{$+0.136$ ($2{\times}10^{-11}$)} \\
$\mathrm{FC}^{\alpha^{*}}$, 2.4m & F-BERT-1M, 2.4m & $+0.031$ (0.18) & \textbf{$+0.051$ ($2{\times}10^{-4}$)} & $-0.007$ (0.57) & \textbf{$+0.350$ ($<10^{-15}$)} \\
$\mathrm{FC}^{\alpha^{*}}$, 2.4m & F-BERT-8M, 2.4m & $+0.035$ (0.07) & \textbf{$+0.036$ ($5{\times}10^{-3}$)} & $+0.004$ (0.70) & \textbf{$+0.159$ ($<10^{-15}$)} \\
F-BERT-1M, 2.4m & Brain-Semantoks, 2.7m & \textbf{$+0.046$ (0.04)} & $-0.051$ ($1{\times}10^{-4}$) & $-0.003$ (0.86) & $-0.014$ (0.48) \\
F-BERT-8M, 2.4m & Brain-Semantoks, 2.7m & $+0.042$ (0.07) & $-0.036$ ($2{\times}10^{-3}$) & $-0.014$ (0.33) & \textbf{$+0.178$ ($<10^{-15}$)} \\
F-BERT-1M, 2.4m & F-BERT-8M, 2.4m & $+0.005$ (0.84) & $-0.015$ (0.23) & $+0.012$ (0.40) & $-0.192$ ($<10^{-15}$) \\
\bottomrule
\end{tabular}
\end{table}

\begin{table}[!htb]
\centering
\small
\setlength{\tabcolsep}{3pt}
\caption{The frozen probe of Table~\ref{tab:bench} on AOMIC-ID1000 (movie watching, $n{=}874$, $10.6$\,min; cognition is Pearson $r$ against the Intelligence Structure Test composite, sex and age are AUC, subject-level CV) and ABIDE-I (autism diagnosis, AUC, site-grouped CV on the all-model intersection, $n{=}158$), with the groups, input columns, bold, underline and asterisk rules of Table~\ref{tab:bench} (paired tests in Table~\ref{tab:paired-aomic}). ``full'' is the whole recording of each cohort; the window rows read the first minutes stated. $^{\S}$As in Table~\ref{tab:bench}.}
\label{tab:cohorts}
\begin{tabular}{@{}lrlrcccc@{}}
\toprule
& \multicolumn{3}{c}{input} & \multicolumn{3}{c}{AOMIC-ID1000} & ABIDE-I \\
\cmidrule(lr){2-4}\cmidrule(lr){5-7}\cmidrule(lr){8-8}
encoder & params & parcels & scan & cognition & sex & age & autism \\
\midrule
\multicolumn{8}{l}{\emph{reads the full scan}} \\
$\mathrm{FC}^{\alpha^{*}}$ (ours) & 79.8k & Sch-400 & full & $\mathbf{0.430 \pm .087}$$^{*}$ & $\mathbf{0.97 \pm .02}$$^{*}$ & $\mathbf{0.55 \pm .04}$ & $\mathbf{0.778 \pm .083}$ \\
FC & 79.8k & Sch-400 & full & $0.368 \pm .095$ & $0.90 \pm .03$ & $0.53 \pm .04$ & $0.751 \pm .092$ \\
\cmidrule(lr){1-8}
F-BERT-8M (ours) & 8.1M & Sch-450 & full & $\mathbf{0.364 \pm .087}$ & $\mathbf{0.92 \pm .03}$$^{*}$ & $0.52 \pm .04$ & $0.671 \pm .014$ \\
F-BERT-1M (ours) & 0.85M & Sch-450 & full & $\underline{0.325 \pm .092}$ & $0.88 \pm .03$ & $\underline{0.53 \pm .04}$ & $\underline{0.707 \pm .047}$ \\
Brain-Semantoks & 63M & Sch-457 & full$^{\S}$ & $0.279 \pm .094$ & $\underline{0.90 \pm .03}$ & $\mathbf{0.54 \pm .04}$ & $\mathbf{0.737 \pm .033}$ \\
BrainMass & 10.2M & Sch-100 & full & $0.303 \pm .105$ & $0.80 \pm .05$ & $0.50 \pm .04$ & $0.670 \pm .062$ \\
\midrule
\multicolumn{8}{l}{\emph{reads a window of the scan}} \\
$\mathrm{FC}^{\alpha^{*}}$ (ours) & 101k & Sch-450 & 2.4m & $\mathbf{0.315 \pm .090}$$^{*}$ & $\mathbf{0.91 \pm .03}$$^{*}$ & $\mathbf{0.54 \pm .04}$$^{*}$ & $\mathbf{0.706 \pm .058}$ \\
FC & 101k & Sch-450 & 2.4m & $0.165 \pm .090$ & $0.85 \pm .04$ & $0.52 \pm .04$ & $0.681 \pm .078$ \\
\cmidrule(lr){1-8}
F-BERT-1M (ours) & 0.85M & Sch-450 & 2.4m & $0.293 \pm .090$ & $0.85 \pm .04$ & $\underline{0.51 \pm .03}$ & $0.595 \pm .039$ \\
F-BERT-8M (ours) & 8.1M & Sch-450 & 2.4m & $\mathbf{0.302 \pm .090}$ & $\mathbf{0.87 \pm .03}$ & $\mathbf{0.52 \pm .04}$ & $\mathbf{0.683 \pm .077}$ \\
Brain-Semantoks & 63M & Sch-457 & 2.7m & $\underline{0.294 \pm .093}$ & $\underline{0.87 \pm .04}$ & $0.49 \pm .03$ & $0.635 \pm .043$ \\
BrainHarmonix & 88M & Sch-400 & 6.4m & $0.017 \pm .102$ & $0.59 \pm .05$ & $0.49 \pm .03$ & $0.428 \pm .046$ \\
BrainLM-111M & 113M & A424 & 2.4m & $0.110 \pm .104$ & $0.68 \pm .05$ & $0.50 \pm .04$ & $\underline{0.660 \pm .037}$ \\
Brain-JEPA & 86M & Sch-450 & 1.9m & $0.036 \pm .095$ & $0.56 \pm .06$ & $0.51 \pm .04$ & $0.560 \pm .037$ \\
BrainLM-650M & 658M & A424 & 2.4m & $0.045 \pm .110$ & $0.54 \pm .06$ & $0.48 \pm .04$ & $0.657 \pm .117$ \\
\bottomrule
\end{tabular}
\end{table}

\begin{table}[!htb]
\centering
\small
\caption{ADHD-200 diagnosis (AUC) under the frozen probe of Table~\ref{tab:bench}, $n{=}298$ over two acquisition sites, site-grouped CV; groups, bold, underline and asterisk rules as in Table~\ref{tab:cohorts} (paired tests in Table~\ref{tab:paired-aomic}). BrainMass, BrainHarmonix and both BrainLM models are absent because their inputs cover only one of the two sites, so no site-grouped fold can be formed.}
\label{tab:adhd}
\begin{tabular}{@{}lrlrc@{}}
\toprule
encoder & params & parcels & scan & ADHD \\
\midrule
\multicolumn{5}{l}{\emph{reads the full scan}} \\
$\mathrm{FC}^{\alpha^{*}}$ (ours) & 79.8k & Sch-400 & full & $\mathbf{0.565 \pm .007}$$^{*}$ \\
FC & 79.8k & Sch-400 & full & $0.560 \pm .009$ \\
\cmidrule(lr){1-5}
F-BERT-8M (ours) & 8.1M & Sch-450 & full & $0.504 \pm .004$ \\
F-BERT-1M (ours) & 0.85M & Sch-450 & full & $\underline{0.523 \pm .020}$ \\
Brain-Semantoks & 63M & Sch-457 & full$^{\S}$ & $\mathbf{0.557 \pm .002}$ \\
\midrule
\multicolumn{5}{l}{\emph{reads a window of the scan}} \\
$\mathrm{FC}^{\alpha^{*}}$ (ours) & 101k & Sch-450 & 2.4m & $0.538 \pm .019$ \\
FC & 101k & Sch-450 & 2.4m & $\mathbf{0.545 \pm .030}$ \\
\cmidrule(lr){1-5}
F-BERT-1M (ours) & 0.85M & Sch-450 & 2.4m & $\mathbf{0.574 \pm .003}$$^{*}$ \\
F-BERT-8M (ours) & 8.1M & Sch-450 & 2.4m & $\underline{0.557 \pm .021}$ \\
Brain-Semantoks & 63M & Sch-457 & 2.7m & $0.537 \pm .013$ \\
Brain-JEPA & 86M & Sch-450 & 1.9m & $0.508 \pm .018$ \\
\bottomrule
\end{tabular}
\end{table}

\begin{table}[!htb]
\centering
\footnotesize\setlength{\tabcolsep}{4pt}
\caption{Paired tests behind Table~\ref{tab:cohorts}, AOMIC-ID1000 cognition ($n{=}874$), as in Table~\ref{tab:paired}: pooled $r$ of the table against the mean of the $200$ per-fold correlations (top), and per pair and target the difference of per-fold means with its corrected resampled $t$-test (bottom): $200$ folds for AOMIC cognition, $100$ for AOMIC sex and age, $30$ site-grouped folds for ABIDE-I ($10$ repetitions of $4$ site folds, those whose held-out site has one class skipped; test fraction $1/4$ in the correction) and $20$ for ADHD-200 ($10$ repetitions over $2$ sites, test fraction $1$). Bold: first encoder significantly ahead ($p<0.05$), the rule that places the asterisks of Table~\ref{tab:cohorts}; n/r where one encoder cannot be scored on ADHD-200 (see Table~\ref{tab:adhd}).}
\label{tab:paired-aomic}
\begin{tabular}{lcc}
\toprule
encoder & pooled $r$ (Table~\ref{tab:cohorts}) & per-fold $r$ \\
\midrule
$\mathrm{FC}^{\alpha^{*}}$ & $0.430$ & $0.422 \pm 0.087$ \\
FC & $0.368$ & $0.366 \pm 0.095$ \\
F-BERT-8M & $0.364$ & $0.366 \pm 0.087$ \\
F-BERT-1M & $0.325$ & $0.329 \pm 0.092$ \\
Brain-Semantoks & $0.279$ & $0.281 \pm 0.094$ \\
BrainMass & $0.303$ & $0.303 \pm 0.105$ \\
$\mathrm{FC}^{\alpha^{*}}$, 2.4m & $0.315$ & $0.313 \pm 0.090$ \\
FC, 2.4m & $0.165$ & $0.195 \pm 0.090$ \\
F-BERT-1M, 2.4m & $0.293$ & $0.294 \pm 0.090$ \\
F-BERT-8M, 2.4m & $0.302$ & $0.303 \pm 0.090$ \\
Brain-Semantoks, 2.7m & $0.294$ & $0.296 \pm 0.093$ \\
BrainLM-111M & $0.110$ & $0.117 \pm 0.104$ \\
BrainHarmonix & $0.017$ & $0.025 \pm 0.102$ \\
Brain-JEPA & $0.036$ & $0.059 \pm 0.095$ \\
BrainLM-650M & $0.045$ & $0.054 \pm 0.110$ \\
\bottomrule
\end{tabular}
\\[6pt]
\scriptsize\setlength{\tabcolsep}{2.5pt}
\resizebox{\linewidth}{!}{\begin{tabular}{@{}llccccc@{}}
\toprule
& & \multicolumn{3}{c}{AOMIC-ID1000} & ABIDE-I & ADHD-200 \\
\cmidrule(lr){3-5}\cmidrule(lr){6-6}\cmidrule(lr){7-7}
first & second & cognition & sex & age & autism & ADHD \\
\midrule
$\mathrm{FC}^{\alpha^{*}}$ & FC & \textbf{$+0.056$ ($2{\times}10^{-3}$)} & \textbf{$+0.072$ ($1{\times}10^{-12}$)} & $+0.011$ (0.26) & $+0.027$ (0.10) & \textbf{$+0.005$ ($2{\times}10^{-3}$)} \\
$\mathrm{FC}^{\alpha^{*}}$, 2.4m & FC, 2.4m & \textbf{$+0.118$ ($3{\times}10^{-9}$)} & \textbf{$+0.066$ ($3{\times}10^{-10}$)} & \textbf{$+0.024$ ($8{\times}10^{-3}$)} & $+0.025$ (0.34) & $-0.007$ (0.56) \\
$\mathrm{FC}^{\alpha^{*}}$ & F-BERT-8M & \textbf{$+0.056$ ($7{\times}10^{-3}$)} & \textbf{$+0.054$ ($1{\times}10^{-11}$)} & \textbf{$+0.036$ ($1{\times}10^{-2}$)} & \textbf{$+0.107$ (0.03)} & \textbf{$+0.061$ ($4{\times}10^{-5}$)} \\
$\mathrm{FC}^{\alpha^{*}}$ & F-BERT-1M & \textbf{$+0.093$ ($6{\times}10^{-5}$)} & \textbf{$+0.094$ ($1{\times}10^{-15}$)} & $+0.017$ (0.24) & \textbf{$+0.071$ ($4{\times}10^{-3}$)} & \textbf{$+0.042$ ($6{\times}10^{-3}$)} \\
FC & F-BERT-8M & $+0.001$ (0.98) & $-0.018$ (0.09) & $+0.025$ (0.11) & $+0.080$ (0.15) & \textbf{$+0.055$ ($4{\times}10^{-4}$)} \\
FC & F-BERT-1M & $+0.037$ (0.18) & $+0.022$ (0.10) & $+0.006$ (0.69) & $+0.044$ (0.20) & \textbf{$+0.037$ ($6{\times}10^{-3}$)} \\
F-BERT-8M & F-BERT-1M & \textbf{$+0.036$ (0.01)} & \textbf{$+0.040$ ($5{\times}10^{-5}$)} & $-0.019$ (0.17) & $-0.036$ (0.16) & $-0.019$ (0.46) \\
F-BERT-1M, 2.4m & F-BERT-8M, 2.4m & $-0.009$ (0.65) & $-0.024$ (0.04) & $-0.009$ (0.57) & $-0.088$ ($6{\times}10^{-4}$) & $+0.018$ (0.48) \\
$\mathrm{FC}^{\alpha^{*}}$, 2.4m & F-BERT-1M, 2.4m & $+0.019$ (0.42) & \textbf{$+0.058$ ($3{\times}10^{-8}$)} & \textbf{$+0.038$ (0.01)} & \textbf{$+0.111$ ($3{\times}10^{-6}$)} & $-0.036$ (0.12) \\
$\mathrm{FC}^{\alpha^{*}}$, 2.4m & F-BERT-8M, 2.4m & $+0.011$ (0.62) & \textbf{$+0.034$ ($9{\times}10^{-4}$)} & \textbf{$+0.029$ (0.04)} & $+0.023$ (0.53) & $-0.019$ ($1{\times}10^{-7}$) \\
F-BERT-1M, 2.4m & FC, 2.4m & \textbf{$+0.099$ ($5{\times}10^{-4}$)} & $+0.008$ (0.59) & $-0.013$ (0.36) & $-0.086$ (0.05) & $+0.030$ (0.40) \\
F-BERT-8M, 2.4m & FC, 2.4m & \textbf{$+0.107$ ($8{\times}10^{-5}$)} & $+0.031$ (0.06) & $-0.005$ (0.75) & $+0.002$ (0.98) & $+0.012$ (0.22) \\
F-BERT-8M & Brain-Semantoks & \textbf{$+0.085$ ($2{\times}10^{-5}$)} & \textbf{$+0.020$ (0.02)} & $-0.024$ (0.07) & $-0.066$ ($3{\times}10^{-3}$) & $-0.052$ ($<10^{-15}$) \\
F-BERT-8M & BrainMass & $+0.063$ (0.05) & \textbf{$+0.121$ ($7{\times}10^{-13}$)} & $+0.013$ (0.38) & $+0.002$ (0.97) & n/r \\
F-BERT-1M & Brain-Semantoks & \textbf{$+0.048$ (0.02)} & $-0.019$ (0.05) & $-0.005$ (0.70) & $-0.030$ ($5{\times}10^{-3}$) & $-0.034$ (0.17) \\
F-BERT-1M & BrainMass & $+0.026$ (0.42) & \textbf{$+0.082$ ($9{\times}10^{-7}$)} & \textbf{$+0.032$ (0.03)} & \textbf{$+0.038$ (0.02)} & n/r \\
F-BERT-8M, 2.4m & BrainHarmonix & \textbf{$+0.278$ ($4{\times}10^{-9}$)} & \textbf{$+0.269$ ($<10^{-15}$)} & $+0.002$ (0.91) & \textbf{$+0.255$ ($4{\times}10^{-4}$)} & n/r \\
F-BERT-8M, 2.4m & BrainLM-111M & \textbf{$+0.185$ ($3{\times}10^{-5}$)} & \textbf{$+0.192$ ($6{\times}10^{-15}$)} & $+0.010$ (0.60) & $+0.023$ (0.48) & n/r \\
F-BERT-8M, 2.4m & Brain-JEPA & \textbf{$+0.244$ ($4{\times}10^{-8}$)} & \textbf{$+0.312$ ($<10^{-15}$)} & $+0.003$ (0.88) & \textbf{$+0.123$ ($2{\times}10^{-3}$)} & $+0.049$ (0.25) \\
F-BERT-8M, 2.4m & BrainLM-650M & \textbf{$+0.248$ ($1{\times}10^{-8}$)} & \textbf{$+0.316$ ($<10^{-15}$)} & $+0.028$ (0.14) & $+0.026$ (0.71) & n/r \\
F-BERT-8M, 2.4m & Brain-Semantoks, 2.7m & $+0.006$ (0.78) & $+0.005$ (0.66) & $+0.024$ (0.09) & \textbf{$+0.048$ (0.01)} & $+0.020$ (0.58) \\
F-BERT-1M, 2.4m & BrainHarmonix & \textbf{$+0.269$ ($3{\times}10^{-9}$)} & \textbf{$+0.245$ ($<10^{-15}$)} & $-0.007$ (0.68) & \textbf{$+0.167$ ($1{\times}10^{-3}$)} & n/r \\
F-BERT-1M, 2.4m & BrainLM-111M & \textbf{$+0.177$ ($5{\times}10^{-5}$)} & \textbf{$+0.169$ ($2{\times}10^{-11}$)} & $+0.001$ (0.94) & $-0.065$ ($9{\times}10^{-7}$) & n/r \\
F-BERT-1M, 2.4m & Brain-JEPA & \textbf{$+0.235$ ($1{\times}10^{-7}$)} & \textbf{$+0.288$ ($<10^{-15}$)} & $-0.006$ (0.72) & \textbf{$+0.035$ (0.03)} & \textbf{$+0.067$ ($6{\times}10^{-4}$)} \\
F-BERT-1M, 2.4m & BrainLM-650M & \textbf{$+0.240$ ($3{\times}10^{-8}$)} & \textbf{$+0.292$ ($<10^{-15}$)} & $+0.019$ (0.35) & $-0.062$ (0.27) & n/r \\
F-BERT-1M, 2.4m & Brain-Semantoks, 2.7m & $-0.002$ (0.92) & $-0.019$ (0.09) & $+0.015$ (0.31) & $-0.040$ ($3{\times}10^{-4}$) & \textbf{$+0.038$ ($2{\times}10^{-3}$)} \\
\bottomrule
\end{tabular}}
\end{table}

\section{Pretraining ablations}
\label{app:pretrain}

\begin{figure}[!htb]
\centering
\includegraphics[width=0.96\linewidth]{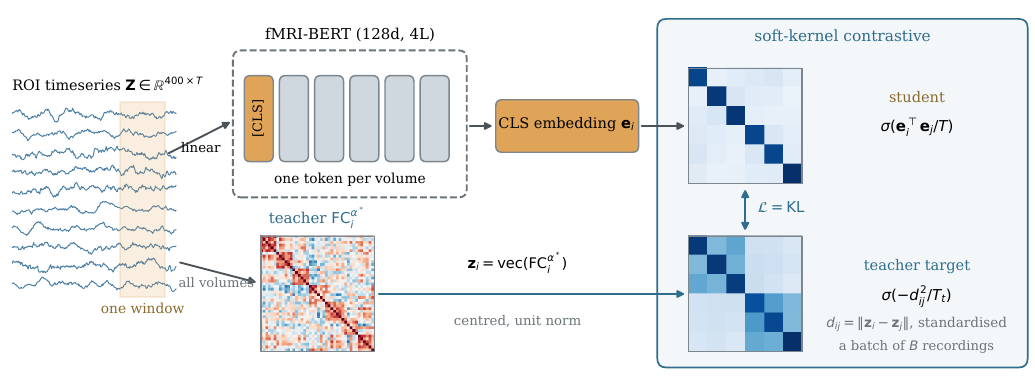}
\caption{The KL objective of Equation~\ref{eq:kl}. The same student and teacher as Figure~\ref{fig:distill}; each recording's row of similarities is turned into a distribution over the other recordings in the batch by a temperature-scaled softmax on both sides, and the two distributions are matched with a KL divergence.}
\label{fig:distill-kl}
\end{figure}

\paragraph{The KL objective.} The alternative to Equation~\ref{eq:cka} matches, for each recording $i$, the student's and the teacher's softmax over the other recordings in the batch. With $\mathbf{e}_i$ the $\ell_2$-normalised embedding, $d_{ij}=\|\mathbf{z}_i-\mathbf{z}_j\|$ and $\tilde d^{\,2}_{ij}$ the squared distances standardised over the batch,
\begin{equation}
s_{ij}=\frac{\mathbf{e}_i^{\top}\mathbf{e}_j}{T},\qquad
t_{ij}=-\frac{\tilde{d}^{\,2}_{ij}}{T_t},\qquad
\mathcal{L}_{\mathrm{KL}}=\frac{1}{B}\sum_i \mathrm{KL}\!\left(\mathrm{softmax}_{j\neq i}(t_{i\cdot})\,\big\|\,\mathrm{softmax}_{j\neq i}(s_{i\cdot})\right),
\label{eq:kl}
\end{equation}
with $T{=}0.07$ and $T_t{=}0.2$. This is momentum contrast \citep{he2020momentum} with the binary labels replaced by the teacher's graded similarities, as in the kernel contrastive loss of \citet{barbano2023contrastive} with the teacher distance in place of a label difference. The standardisation is required: over $101{,}025$ edges the pairwise distances concentrate so strongly that the softmax of the unstandardised distances has $99.7\%$ of the entropy of a uniform distribution. Because the teacher vectors are centred and unit-norm, $d_{ij}^2=2-2r_{ij}$ with $r_{ij}$ the correlation between the two connectomes' edges, so both objectives see the same teacher kernel; they differ in how the match is scored, a per-anchor divergence over neighbours against a global alignment of the two Gram matrices (Figure~\ref{fig:distill-kl}).

\paragraph{Optimisation.} Every encoder is trained with AdamW \citep{loshchilov2019decoupled} (weight decay $10^{-4}$), a linear warm-up of $100$ steps followed by cosine decay over $300{,}000$ steps, gradient-norm clipping at $1$, mixed precision, and seed $0$. The CKA runs use batch $1{,}024$ and learning rate $8.5\times10^{-4}$ at both sizes; the KL runs use batch $128$ and learning rate $3\times10^{-4}$ ($1$M) or $10^{-4}$ with a $2{,}000$-step warm-up ($8$M). Windows are $80$ timepoints drawn uniformly over the corpus; the teacher vector of a window is that of its full recording. A checkpoint is written every $2{,}000$ steps and the reported one is selected on the downstream probe, at step $56{,}000$ for the $1$M encoder and $12{,}000$ for the $8$M; the selection optimism measured by a family-split hold-out is about $0.03$ in $r$. The released code, configurations and corpus reproduce the runs.

\paragraph{Objective and batch size.} Table~\ref{tab:objbatch} trains the $1$M encoder with each objective at several batch sizes, with the learning rate set per run (configurations in the released code), and probes the cognitive composite. The KL objective degrades monotonically as the batch grows, since each row's softmax gets more negatives and a harder, temperature-dependent target; the CKA objective improves, since a larger batch is a better estimate of the two Gram matrices. The best CKA model beats the best KL model on shared folds ($+0.050$, $p_{\mathrm{NB}}=0.012$, ahead in $161$ of $200$ folds).

\begin{table}[!htb]
\centering
\small
\caption{Objective against batch size, $1$M encoder, HCP-YA cognitive composite, windowed protocol ($144$\,s windows averaged per subject, linear-kernel ridge, $20\times10$ family-aware CV), best checkpoint of each run.}
\label{tab:objbatch}
\begin{tabular}{lccc}
\toprule
objective & $B=128$ & $B=256$ & $B=1024$ \\
\midrule
KL (Eq.~\ref{eq:kl}) & $0.461 \pm 0.091$ & $0.442 \pm 0.088$ & $0.413 \pm 0.095$ \\
CKA (Eq.~\ref{eq:cka}) & $0.490 \pm 0.090$ & not run & $\mathbf{0.510 \pm 0.086}$ \\
\bottomrule
\end{tabular}
\end{table}

\paragraph{Scale.} Table~\ref{tab:scale} repeats the comparison at two encoder sizes. The CKA objective gains with the tenfold increase in parameters and reaches its best score in fewer steps ($12$k against $56$k); the KL objective gains little. Under the continuous-sequence protocol of Table~\ref{tab:bench} the ordering is the same, and the $8$M KL model falls below the $1$M one.

\begin{table}[!htb]
\centering
\small
\caption{Scaling, HCP-YA cognitive composite. Windowed protocol as in Table~\ref{tab:objbatch}, and the whole recording read as one sequence as in Table~\ref{tab:bench}.}
\label{tab:scale}
\begin{tabular}{llcc}
\toprule
objective & protocol & $1$M ($128$d, $4$L) & $8$M ($384$d, $8$L) \\
\midrule
CKA & windowed & $0.510 \pm 0.086$ & $0.541 \pm 0.083$ \\
KL & windowed & $0.461 \pm 0.091$ & $0.486 \pm 0.088$ \\
CKA & one sequence & $0.488 \pm 0.090$ & $0.502 \pm 0.083$ \\
KL & one sequence & $0.453 \pm 0.089$ & $0.426 \pm 0.105$ \\
\bottomrule
\end{tabular}
\end{table}

\begin{figure}[!htb]
\centering
\includegraphics[width=\linewidth]{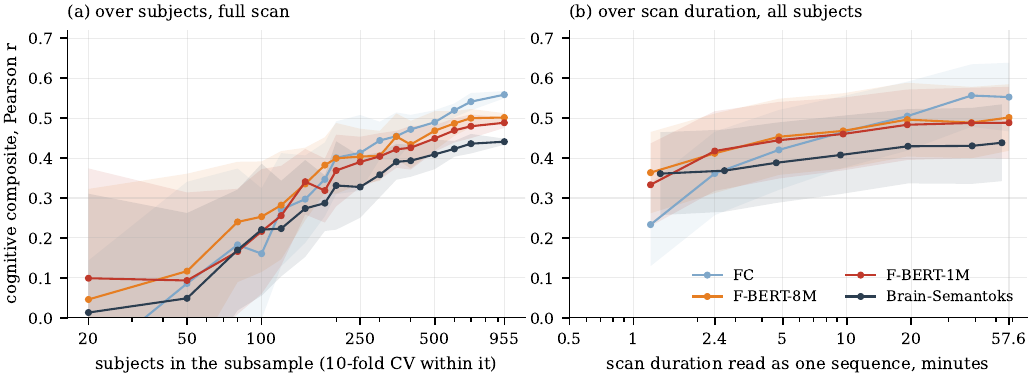}
\caption{Cognitive composite on HCP-YA, linear-kernel ridge probe of Table~\ref{tab:bench}, pooled out-of-fold $r$. \textbf{(a)} Against the number of subjects in a random subsample, full scan, $10$-fold family-aware CV within the subsample; points are the mean over $20$ subsamples and bands their standard deviation. \textbf{(b)} Against the duration read from the start of each recording, $20\times10$ family-aware CV; bands are the standard deviation over the $200$ folds. The encoder points at $2.4$ and $57.6$\,min are the cells of Table~\ref{tab:bench}. FC is at Schaefer-400 at every duration, whereas its $2.4$\,min row in Table~\ref{tab:bench} is at Sch-450, the encoders' input; subjects shorter than a duration are dropped from it ($939$ remain at $57.6$\,min). Brain-Semantoks reads at $\mathrm{TR}=2$\,s and is shown at its own durations.}
\label{fig:duration}
\end{figure}

\paragraph{Subjects and scan duration.} Figure~\ref{fig:duration} gives the cognitive composite against the number of subjects and against the duration of scan read, under the probe and the features of Table~\ref{tab:bench}. In (a) a random subsample of the $955$ subjects is drawn and the probe is run within it; the $955$ column is Table~\ref{tab:bench}. In (b) every representation is computed from the first minutes of each recording, the encoders reading them as one continuous sequence; Brain-Semantoks, whose positions are also sinusoidal, was rebuilt at each length with its released weights. The $8$M model is above raw FC up to $180$ subjects, within $0.01$ of it at $200$ and $250$, and FC is ahead from $300$; over duration both of our models are above raw FC up to $4.8$\,min and FC is ahead from $19.2$\,min. For the $1$M model, averaging the embeddings of consecutive $144$\,s windows instead of reading one long sequence gives $0.49$ to $0.51$ at every window length ($\pm0.09$), so what determines the score is the amount of scan seen; the $8$M model scores higher windowed (Table~\ref{tab:scale}), and Table~\ref{tab:bench} keeps one protocol for both sizes.

\begin{table}[!htb]
\centering
\caption{Gain of F-BERT-1M over raw FC on the cognitive composite by scan length (rows) and number of subjects (columns), HCP-YA Sch-450: mean $\pm$ std of the paired per-fold difference under the probe of Table~\ref{tab:bench}; $^{*}$ one-sided $p_{\mathrm{NB}}<0.05$ in favour of the encoder. Green is a gain, red a loss. F-BERT-8M in Table~\ref{tab:grid8m}, surfaces in Figure~\ref{fig:grid3d}.}
\label{tab:gridfull}
\scriptsize\setlength{\tabcolsep}{1.6pt}
\begin{tabular}{@{}lccccccc@{}}
\toprule
& \multicolumn{7}{c}{subjects} \\
\cmidrule(lr){2-8}
min & $100$ & $150$ & $250$ & $350$ & $500$ & $700$ & all \\
\midrule
1.2 & \cellcolor{green!39}$+0.17\pm0.44$ & \cellcolor{green!33}$+0.15\pm0.35$ & \cellcolor{green!23}$+0.10\pm0.25$ & \cellcolor{green!27}$+0.12\pm0.23$ & \cellcolor{green!20}$+0.09\pm0.18$ & \cellcolor{green!20}$+0.09\pm0.14$$^{*}$ & \cellcolor{green!19}$+0.09\pm0.11$$^{*}$ \\
2.4 & \cellcolor{green!45}$+0.21\pm0.39$ & \cellcolor{green!40}$+0.18\pm0.34$ & \cellcolor{green!29}$+0.13\pm0.20$$^{*}$ & \cellcolor{green!34}$+0.15\pm0.20$$^{*}$ & \cellcolor{green!19}$+0.09\pm0.14$$^{*}$ & \cellcolor{green!18}$+0.08\pm0.11$$^{*}$ & \cellcolor{green!12}$+0.06\pm0.09$$^{*}$ \\
4.8 & \cellcolor{green!45}$+0.22\pm0.37$$^{*}$ & \cellcolor{green!42}$+0.19\pm0.32$$^{*}$ & \cellcolor{green!26}$+0.12\pm0.22$ & \cellcolor{green!19}$+0.09\pm0.17$ & \cellcolor{green!11}$+0.05\pm0.12$ & \cellcolor{green!7}$+0.03\pm0.09$ & \cellcolor{green!5}$+0.03\pm0.08$ \\
9.6 & \cellcolor{green!45}$+0.22\pm0.38$$^{*}$ & \cellcolor{green!33}$+0.15\pm0.29$ & \cellcolor{green!19}$+0.09\pm0.19$ & \cellcolor{green!12}$+0.06\pm0.16$ & \cellcolor{green!5}$+0.02\pm0.11$ & $+0.00\pm0.08$ & $-0.00\pm0.07$ \\
19.2 & \cellcolor{green!31}$+0.14\pm0.40$ & \cellcolor{green!19}$+0.09\pm0.25$ & \cellcolor{green!13}$+0.06\pm0.17$ & $+0.02\pm0.14$ & $+0.01\pm0.11$ & $-0.01\pm0.08$ & $-0.02\pm0.06$ \\
38.4 & \cellcolor{green!23}$+0.11\pm0.42$ & \cellcolor{green!10}$+0.05\pm0.28$ & $+0.02\pm0.16$ & \cellcolor{red!6}$-0.03\pm0.13$ & \cellcolor{red!7}$-0.03\pm0.11$ & \cellcolor{red!11}$-0.05\pm0.07$ & \cellcolor{red!13}$-0.06\pm0.06$ \\
57.6 & \cellcolor{green!7}$+0.03\pm0.32$ & $+0.01\pm0.24$ & $-0.01\pm0.18$ & $-0.01\pm0.15$ & \cellcolor{red!11}$-0.05\pm0.10$ & \cellcolor{red!12}$-0.06\pm0.08$ & \cellcolor{red!15}$-0.07\pm0.06$ \\
\bottomrule
\end{tabular}
\end{table}

\begin{table}[!htb]
\centering
\caption{As Table~\ref{tab:gridfull}, for F-BERT-8M minus raw FC.}
\label{tab:grid8m}
\scriptsize\setlength{\tabcolsep}{1.6pt}
\begin{tabular}{@{}lccccccc@{}}
\toprule
& \multicolumn{7}{c}{subjects} \\
\cmidrule(lr){2-8}
min & $100$ & $150$ & $250$ & $350$ & $500$ & $700$ & all \\
\midrule
1.2 & \cellcolor{green!41}$+0.18\pm0.41$ & \cellcolor{green!26}$+0.12\pm0.36$ & \cellcolor{green!19}$+0.09\pm0.24$ & \cellcolor{green!28}$+0.12\pm0.21$$^{*}$ & \cellcolor{green!21}$+0.10\pm0.17$$^{*}$ & \cellcolor{green!26}$+0.12\pm0.14$$^{*}$ & \cellcolor{green!25}$+0.12\pm0.10$$^{*}$ \\
2.4 & \cellcolor{green!45}$+0.21\pm0.39$ & \cellcolor{green!34}$+0.15\pm0.33$ & \cellcolor{green!23}$+0.10\pm0.20$ & \cellcolor{green!26}$+0.12\pm0.18$$^{*}$ & \cellcolor{green!13}$+0.06\pm0.14$ & \cellcolor{green!14}$+0.07\pm0.11$$^{*}$ & \cellcolor{green!11}$+0.05\pm0.08$$^{*}$ \\
4.8 & \cellcolor{green!43}$+0.20\pm0.38$ & \cellcolor{green!39}$+0.18\pm0.30$$^{*}$ & \cellcolor{green!23}$+0.10\pm0.20$ & \cellcolor{green!15}$+0.07\pm0.17$ & \cellcolor{green!8}$+0.04\pm0.13$ & \cellcolor{green!7}$+0.04\pm0.09$ & \cellcolor{green!7}$+0.03\pm0.06$ \\
9.6 & \cellcolor{green!44}$+0.20\pm0.37$ & \cellcolor{green!30}$+0.13\pm0.29$ & \cellcolor{green!13}$+0.06\pm0.18$ & \cellcolor{green!10}$+0.05\pm0.16$ & $+0.01\pm0.11$ & $+0.00\pm0.08$ & $+0.00\pm0.06$ \\
19.2 & \cellcolor{green!23}$+0.11\pm0.38$ & \cellcolor{green!24}$+0.11\pm0.24$ & \cellcolor{green!11}$+0.05\pm0.17$ & $+0.02\pm0.14$ & $-0.00\pm0.10$ & $-0.01\pm0.08$ & $-0.01\pm0.06$ \\
38.4 & \cellcolor{green!23}$+0.10\pm0.39$ & \cellcolor{green!18}$+0.08\pm0.24$ & $+0.02\pm0.16$ & \cellcolor{red!5}$-0.02\pm0.13$ & \cellcolor{red!6}$-0.03\pm0.09$ & \cellcolor{red!8}$-0.04\pm0.07$ & \cellcolor{red!9}$-0.04\pm0.06$ \\
57.6 & \cellcolor{green!6}$+0.03\pm0.30$ & $+0.01\pm0.25$ & $+0.01\pm0.18$ & $+0.00\pm0.15$ & $-0.02\pm0.10$ & \cellcolor{red!5}$-0.02\pm0.08$ & \cellcolor{red!8}$-0.04\pm0.06$ \\
\bottomrule
\end{tabular}
\end{table}

\begin{table}[!htb]
\centering
\caption{Gain of each encoder over raw FC in CoRR identification accuracy by scan length (rows) and number of subjects (columns), Sch-450: mean $\pm$ std of the difference over $20$ random subject subsets shared by both; $^{*}$ one-sided paired $t$-test $p<0.05$ in favour of the encoder. Scans shorter than the row's length are dropped, so the $6$-minute row has $262$ subjects; -- where the column exceeds them.}
\label{tab:gridfp}
\scriptsize\setlength{\tabcolsep}{2pt}
\begin{tabular}{@{}lcccccc@{}}
\toprule
\multicolumn{7}{l}{\emph{F-BERT-1M minus FC}} \\
& \multicolumn{6}{c}{subjects} \\
\cmidrule(lr){2-7}
min & $100$ & $150$ & $200$ & $250$ & $300$ & all \\
\midrule
0.7 & \cellcolor{red!37}$-0.17\pm0.04$ & \cellcolor{red!35}$-0.16\pm0.03$ & \cellcolor{red!32}$-0.15\pm0.02$ & \cellcolor{red!31}$-0.14\pm0.02$ & \cellcolor{red!32}$-0.15\pm0.01$ & \cellcolor{red!31}$-0.14\pm0.00$ \\
1.3 & \cellcolor{red!19}$-0.08\pm0.04$ & \cellcolor{red!19}$-0.09\pm0.03$ & \cellcolor{red!20}$-0.09\pm0.01$ & \cellcolor{red!20}$-0.09\pm0.02$ & \cellcolor{red!22}$-0.10\pm0.02$ & \cellcolor{red!22}$-0.10\pm0.00$ \\
2 & \cellcolor{red!12}$-0.06\pm0.05$ & \cellcolor{red!14}$-0.07\pm0.04$ & \cellcolor{red!16}$-0.07\pm0.02$ & \cellcolor{red!16}$-0.07\pm0.02$ & \cellcolor{red!18}$-0.08\pm0.01$ & \cellcolor{red!19}$-0.09\pm0.00$ \\
2.7 & \cellcolor{red!12}$-0.06\pm0.04$ & \cellcolor{red!14}$-0.06\pm0.04$ & \cellcolor{red!15}$-0.07\pm0.03$ & \cellcolor{red!16}$-0.07\pm0.02$ & \cellcolor{red!16}$-0.07\pm0.01$ & \cellcolor{red!18}$-0.08\pm0.00$ \\
4 & \cellcolor{red!16}$-0.07\pm0.04$ & \cellcolor{red!22}$-0.10\pm0.02$ & \cellcolor{red!21}$-0.09\pm0.02$ & \cellcolor{red!21}$-0.10\pm0.02$ & \cellcolor{red!21}$-0.10\pm0.02$ & \cellcolor{red!20}$-0.09\pm0.00$ \\
6 & $-0.02\pm0.03$ & $-0.02\pm0.02$ & \cellcolor{red!6}$-0.03\pm0.01$ & \cellcolor{red!7}$-0.03\pm0.01$ & -- & \cellcolor{red!6}$-0.03\pm0.00$ \\
full & \cellcolor{red!13}$-0.06\pm0.03$ & \cellcolor{red!16}$-0.07\pm0.02$ & \cellcolor{red!18}$-0.08\pm0.01$ & \cellcolor{red!18}$-0.08\pm0.01$ & \cellcolor{red!19}$-0.09\pm0.01$ & \cellcolor{red!20}$-0.09\pm0.00$ \\
\midrule
\multicolumn{7}{l}{\emph{F-BERT-8M minus FC}} \\
& \multicolumn{6}{c}{subjects} \\
\cmidrule(lr){2-7}
min & $100$ & $150$ & $200$ & $250$ & $300$ & all \\
\midrule
0.7 & \cellcolor{green!17}$+0.08\pm0.04$$^{*}$ & \cellcolor{green!15}$+0.07\pm0.03$$^{*}$ & \cellcolor{green!16}$+0.07\pm0.02$$^{*}$ & \cellcolor{green!14}$+0.06\pm0.02$$^{*}$ & \cellcolor{green!12}$+0.05\pm0.02$$^{*}$ & \cellcolor{green!9}$+0.04\pm0.00$$^{*}$ \\
1.3 & \cellcolor{green!43}$+0.20\pm0.04$$^{*}$ & \cellcolor{green!41}$+0.18\pm0.02$$^{*}$ & \cellcolor{green!42}$+0.19\pm0.02$$^{*}$ & \cellcolor{green!43}$+0.19\pm0.03$$^{*}$ & \cellcolor{green!41}$+0.19\pm0.01$$^{*}$ & \cellcolor{green!41}$+0.19\pm0.00$$^{*}$ \\
2 & \cellcolor{green!31}$+0.14\pm0.04$$^{*}$ & \cellcolor{green!31}$+0.14\pm0.03$$^{*}$ & \cellcolor{green!32}$+0.15\pm0.02$$^{*}$ & \cellcolor{green!33}$+0.15\pm0.02$$^{*}$ & \cellcolor{green!32}$+0.15\pm0.02$$^{*}$ & \cellcolor{green!33}$+0.15\pm0.00$$^{*}$ \\
2.7 & \cellcolor{green!24}$+0.11\pm0.03$$^{*}$ & \cellcolor{green!24}$+0.11\pm0.04$$^{*}$ & \cellcolor{green!23}$+0.11\pm0.02$$^{*}$ & \cellcolor{green!24}$+0.11\pm0.02$$^{*}$ & \cellcolor{green!23}$+0.11\pm0.01$$^{*}$ & \cellcolor{green!25}$+0.11\pm0.00$$^{*}$ \\
4 & \cellcolor{green!14}$+0.06\pm0.03$$^{*}$ & \cellcolor{green!12}$+0.06\pm0.02$$^{*}$ & \cellcolor{green!15}$+0.07\pm0.02$$^{*}$ & \cellcolor{green!12}$+0.06\pm0.02$$^{*}$ & \cellcolor{green!14}$+0.06\pm0.01$$^{*}$ & \cellcolor{green!13}$+0.06\pm0.00$$^{*}$ \\
6 & \cellcolor{green!13}$+0.06\pm0.03$$^{*}$ & \cellcolor{green!16}$+0.07\pm0.02$$^{*}$ & \cellcolor{green!16}$+0.07\pm0.01$$^{*}$ & \cellcolor{green!16}$+0.07\pm0.01$$^{*}$ & -- & \cellcolor{green!16}$+0.07\pm0.00$$^{*}$ \\
full & \cellcolor{green!15}$+0.07\pm0.03$$^{*}$ & \cellcolor{green!18}$+0.08\pm0.02$$^{*}$ & \cellcolor{green!17}$+0.08\pm0.02$$^{*}$ & \cellcolor{green!18}$+0.08\pm0.01$$^{*}$ & \cellcolor{green!17}$+0.08\pm0.01$$^{*}$ & \cellcolor{green!17}$+0.08\pm0.00$$^{*}$ \\
\bottomrule
\end{tabular}
\end{table}

\begin{figure}[!htb]
\centering
\includegraphics[width=\linewidth]{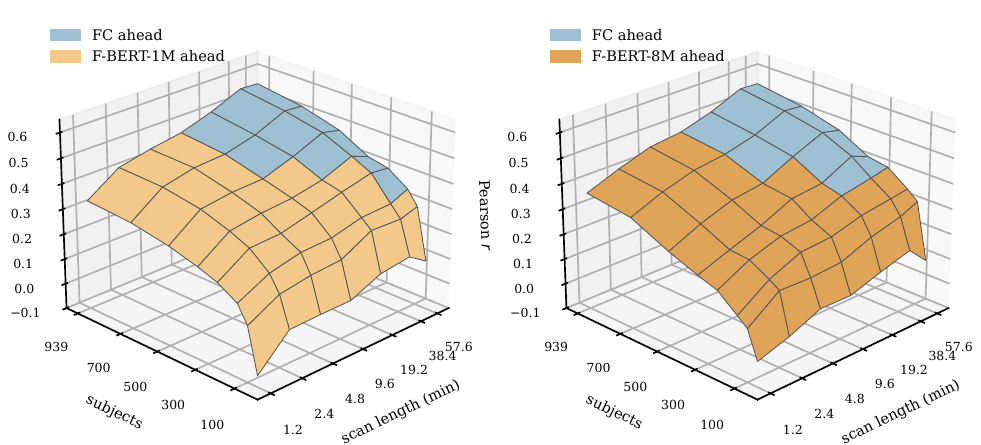}
\caption{Cognitive composite (pooled out-of-fold $r$, probe of Table~\ref{tab:bench}) over scan length and number of subjects, HCP-YA Sch-450, shown as the higher of raw FC and the encoder at each cell, coloured by which one is ahead; Tables~\ref{tab:gridfull} and \ref{tab:grid8m} give the paired differences.}
\label{fig:grid3d}
\end{figure}

\section{Distillation details}
\label{app:distill}

\paragraph{Why the distance teacher is the correlation kernel.} The KL objective of Equation~\ref{eq:kl} grades each pair by $d_{ij}=\|\mathbf{z}_i-\mathbf{z}_j\|$. Because $\mathbf{z}$ is centred on its own mean and scaled to unit norm, $\mathbf{z}_i^{\top}\mathbf{z}_j$ is the Pearson correlation $r_{ij}$ between the two connectomes' edges, and
\begin{equation}
d_{ij}^2=\|\mathbf{z}_i\|^2+\|\mathbf{z}_j\|^2-2\,\mathbf{z}_i^{\top}\mathbf{z}_j=2-2r_{ij}.
\end{equation}
The loss uses $-d_{ij}^2/T_t$, so a more correlated pair receives a larger teacher logit. Standardising makes the correspondence exact rather than merely affine: $\sigma_{d^2}=2\sigma_r$ gives $-(d^2-\mu_{d^2})/\sigma_{d^2}=(r-\mu_r)/\sigma_r$, and the constant that remains is shared across a row and cancels in the softmax. On teacher vectors built by the released precompute the two teacher distributions agree to $4\times10^{-14}$ in log-probability ($\mathrm{KL}=2\times10^{-17}$), so training against the distance and training against the correlation kernel are the same objective. The implication runs one way only: this holds because the vectors are unit-norm, and a distance teacher on unnormalised vectors would not reduce to a correlation.

\paragraph{No positive pair; augmentation optional.} A contrastive objective normally builds its positive pair by augmenting a sample. Here the supervision comes from the teacher's weighting of every other recording in the batch, so no positive pair is needed, and the released runs use one view per recording. Augmentations are admissible as long as they leave the teacher unchanged: any transformation of the timeseries that preserves the FC matrix yields a second view with the same $\mathrm{FC}^{\alpha^{*}}$. One such transformation is implemented in the released code, the Fourier phase surrogate that multiplies the spectrum of every region by one shared random phase, which preserves every cross-spectrum and therefore the FC matrix. Trained with it under the KL objective (batch $64$, distance teacher), the best checkpoint scores $0.436\pm0.093$ against $0.433\pm0.090$ without it on the windowed protocol, and at a matched step the paired difference is $+0.042$ ($p_{\mathrm{NB}}=0.15$, $143$ of $200$ folds), so it is not used in the reported models. As $T_t\to0$ the teacher distribution concentrates on the single most similar recording and the loss reduces to a hard contrastive loss whose positive is the nearest other recording rather than a second view of the same one.

\subsection{AOMIC parcellation sweep}
\label{app:aomicparc}

Table~\ref{tab:main} reports AOMIC at the four Schaefer resolutions prepared by
the original pipeline. To test the exponent against a wider range of
parcellations we re-parcellated AOMIC from the fMRIPrep derivatives at six
atlases in one pass, including A424, which is not a Schaefer parcellation.

These rows are internally consistent but are \emph{not} comparable with
Table~\ref{tab:main}: re-parcellating with our own confound set and band
reproduces the original Schaefer-400 timeseries only to a median per-region
correlation of $0.60$, and the resulting FC edge vectors to $0.77$. The
Schaefer-400 row below is therefore the bridge, and it is the pipeline
difference rather than the atlas that separates it from the corresponding row
of Table~\ref{tab:main}. What carries across is the sign and the size of the
gain, on every atlas including one from a different family.

\begin{table}[!htb]
\centering
\small
\caption{FC$^{\alpha^{*}}$ against FC on AOMIC-ID1000 under a single
re-parcellation pipeline ($n{=}879$ subjects with the composite under that pipeline), correlation-kernel KRR, $20\times10$ subject-level CV, mean $\pm$ std over the $200$ folds.}
\begin{tabular}{lccccc}
\toprule
parcellation & FC ($\alpha{=}1$) & $\mathrm{FC}^{\alpha^{*}}$ & $\Delta$ & $p_{\mathrm{NB}}$ & folds won \\
\midrule
Schaefer-400 & $0.302 \pm 0.085$ & $0.373 \pm 0.085$ & $+0.071$ & $8{\times}10^{-5}$ & 181/200 \\
Schaefer-500 & $0.309 \pm 0.081$ & $0.390 \pm 0.080$ & $+0.081$ & $4{\times}10^{-6}$ & 191/200 \\
Schaefer-600 & $0.314 \pm 0.082$ & $0.405 \pm 0.081$ & $+0.091$ & $1{\times}10^{-7}$ & 197/200 \\
Schaefer-800 & $0.312 \pm 0.083$ & $0.404 \pm 0.082$ & $+0.092$ & $3{\times}10^{-8}$ & 197/200 \\
Schaefer-1000 & $0.312 \pm 0.085$ & $0.401 \pm 0.085$ & $+0.089$ & $2{\times}10^{-8}$ & 197/200 \\
A424 & $0.287 \pm 0.086$ & $0.378 \pm 0.084$ & $+0.092$ & $1{\times}10^{-6}$ & 191/200 \\
\bottomrule
\end{tabular}
\end{table}

\subsection{ADHD-200}
\label{app:adhd}

ADHD-200 leaves $298$ labelled scans across two acquisition sites after the all-encoder intersection, each held out in turn; the remaining encoders cover only $119$ of those scans, all from one site, so they cannot be scored. Every scorable representation is close to chance (Table~\ref{tab:adhd}), and we report it as a null result rather than a column of Table~\ref{tab:bench}.

\end{document}